\documentclass[lettersize,journal]{IEEEtran}
\usepackage{amsmath,amsfonts}
\usepackage{array}
\usepackage[caption=false,font=normalsize,labelfont=sf,textfont=sf]{subfig}
\usepackage{textcomp}
\usepackage{stfloats}
\usepackage{url}
\usepackage{verbatim}
\usepackage{graphicx}
\usepackage{cite}

\usepackage{times}
\usepackage{color}
\usepackage{multirow}
\usepackage{amssymb}
\usepackage{amsthm}
\usepackage{diffcoeff}
\usepackage{epstopdf}
\usepackage{rotating}
\usepackage{bbm}
\usepackage{latexsym}
\usepackage{wrapfig}
\usepackage{booktabs}

\usepackage{algorithm}
\usepackage{algorithmicx}
\usepackage{algpseudocode}
\floatname{algorithm}{Algorithm} 
\usepackage{algorithm,algpseudocode,float}
\usepackage{lipsum}


\usepackage{balance}
\begin{document}

\title{Learning Network Representations with Disentangled Graph Auto-Encoder}
\author{Di Fan and Chuanhou Gao,~\IEEEmembership{Senior Member,~IEEE}
\thanks{This work was funded by the National Nature Science Foundation of China under Grant No. 12320101001 and 12071428. \textit{(Corresponding author: Chuanhou Gao.)}}
\thanks{Di Fan and Chuanhou Gao are with the School of Mathematical Sciences, Zhejiang University, Hangzhou 310030, China (e-mail:
fandi@zju.edu.cn; gaochou@zju.edu.cn).}}
\markboth{Journal of \LaTeX\ Class Files,~Vol.~14, No.~8, August~2021}%
{Shell \MakeLowercase{\textit{et al.}}: A Sample Article Using IEEEtran.cls for IEEE Journals}

\IEEEpubid{0000--0000/00\$00.00~\copyright~2021 IEEE}

\maketitle
\begin{abstract}
The (variational) graph auto-encoder is widely used to learn representations for graph-structured data. However, the formation of real-world graphs is a complicated and heterogeneous process influenced by latent factors. Existing encoders are fundamentally holistic, neglecting the entanglement of latent factors. This reduces the effectiveness of graph analysis tasks, while also making it more difficult to explain the learned representations. As a result, learning disentangled graph representations with the (variational) graph auto-encoder poses significant challenges and remains largely unexplored in the current research. In this paper, we introduce the Disentangled Graph Auto-Encoder (DGA) and the Disentangled Variational Graph Auto-Encoder (DVGA) to learn disentangled representations. Specifically, we first design a disentangled graph convolutional network with multi-channel message-passing layers to serve as the encoder. This allows each channel to aggregate information about each latent factor.
The disentangled variational graph auto-encoder's expressive capability is then enhanced by applying a component-wise flow to each channel. In addition, we construct a factor-wise decoder that takes into account the characteristics of disentangled representations. We improve the independence of representations by imposing independence constraints on the mapping channels for distinct latent factors. Empirical experiments on both synthetic and real-world datasets demonstrate the superiority of our proposed method compared to several state-of-the-art baselines.
\end{abstract}
\section{Introduction}
\IEEEPARstart{G}{raph}-structured data has become exceptionally prevalent in the era of big data, spanning across various domains, including social networks, co-authorship networks, biological networks and traffic networks, etc. Represented in the form of graphs, these data illustrate the complex interconnections between entities, reflecting intricate and extensive networks of relationships. 

The field of graph analysis includes many significant tasks. Link prediction \cite{Loehlin2004latent,wang2020neighborhood,Son2023expert} is a well-known task in which missing or potential connections in a network are predicted. For example, in social networks, link prediction facilitates precise  targeting for target marketing, hence informing marketing strategies. Biological networks use it to forecast associations between molecules, whereas terrorist networks use it to identify individuals who arouse suspicion. Additional prevalent graph analysis tasks include node clustering \cite{Wang2021mgae,Tsitsulin2023graph} and node classification \cite{Bhagat2011node,Xiao2022graph,Luan2023when}. Node clustering is the process of grouping nodes in a graph with the goal of ensuring that nodes inside each cluster are closely related. 
Node classification involves assigning predefined labels to nodes based on their attributes, and this work is essential for gaining insights into the specific characteristics of each node in a graph. These tasks also have wide-ranging applications in areas such as social networks, biology, network security 
and medical analysis.
\IEEEpubidadjcol

\begin{figure}[!t]
\begin{center}
\includegraphics[width=0.3\textwidth]{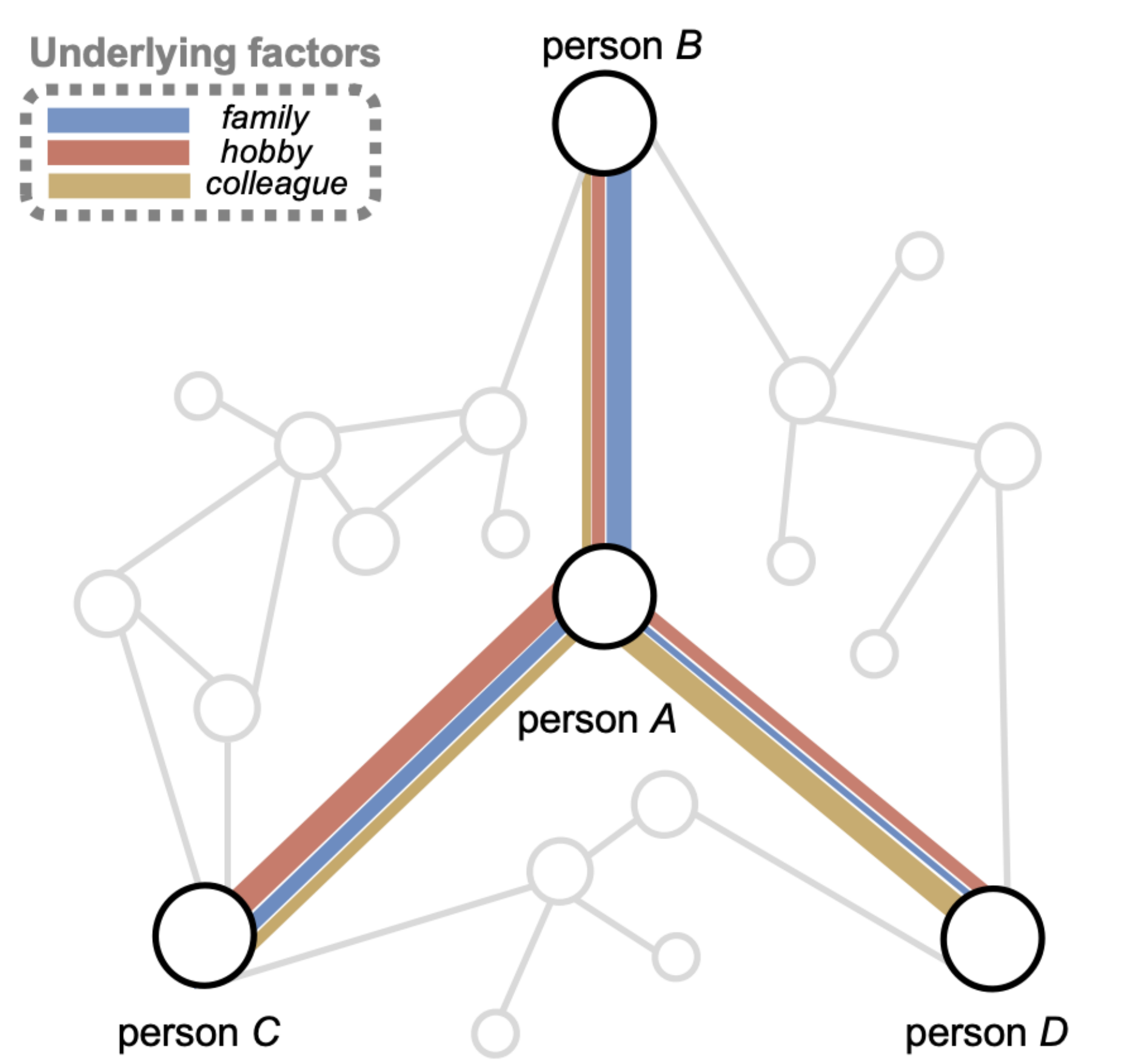}
\end{center}
\caption{An illustrative social network example that inspired our work. We consider using lines of different colors to represent different underlying factors that contribute to the relationship between two persons. The thickness of the lines is directly proportional to the frequency of interaction due to these factors.}
\label{Fig:social network}
\end{figure}

However, due to the extensive and highly sparse nature of graphs, graph analysis tasks are often challenging. Therefore, there has been a growing interest in improving the learning of graph representations \cite{Wang2016structural,Wang2017community,liu2023hgber}. 
The tasks of learning node representations serve as the basis for the previously mentioned network analysis tasks. An effective approach for learning node representations without supervision is the Variational Graph Auto-Encoder (VGAE) and the Graph Auto-Encoder (GAE) \cite{Kipf2016variational}. They all consist of an encoder responsible for learning latent representations and a decoder tasked with reconstructing the graph structure data using the learned representations. Their primary objective is to successfully complete the task of link prediction. Consequently, the decoder is relatively simple, depending only on the inner product of the latent representations of the nodes. 
The encoder is based solely on the original Graph Convolutional Networks (GCN) \cite{Kipf2016semi}. 
These methods take a holistic approach and treat the graph as a perceptual whole \cite{Kipf2016variational,Hasanzadeh2019Semi,Grover2019graphite,Khan2021epitomic,Shi2020effective,Sun2021interpretable,Li2023variational}. They ignore subtle differences between various relationships in the graph. In fact, graph formation in the real world is a complex, heterogeneous process driven by latent factors. Figure \ref{Fig:social network} illustrates the concept of disentanglement. It is evident that the connections between the four persons are established based on several underlying factors. Moreover, their relationships exhibit distinct emphases. For instance, the relationship between person $A$ and person $B$ is mainly due to familial connections, whereas the connection between person $A$ and person $C$ is predominantly based on their shared hobby. Additionally, it is important to acknowledge that relationships between any pair of individuals are influenced by multiple factors, while the emphasis placed on these aspects may differ. Without this precise differentiation in relationships, representing nodes only from binary network topologies becomes problematic. This not only hinders the interpretability of learned representations but also results in diminished performance on subsequent tasks. Hence, the intricate nature of the problem forces the encoder to disentangle factors, a crucial aspect that current holistic methods have overlooked.

In this paper, we propose learning disentangled node representations. Disentangled representation learning, which seeks to identify different underlying factors in factorized representations, has shown many advantages \cite{higgins2018towards}. However, it encounters three problems when applied to graph auto-encoder models. (1) Designing a graph encoder for disentangled representations requires careful planning to guarantee it can effectively infer latent factors with enough expressiveness. (2) Designing a decoder for disentangled representations requires taking into account the factor information in the connections between nodes in order to enhance the accuracy of link prediction. (3) To improve the quality of disentangled representations, a training strategy should be designed to ensure statistical independence between distinct parts of representations that correspond to different latent factors.

To tackle the aforementioned challenges, we introduce two innovative auto-encoder models for learning disentangled graph representations, that is, Disentangled Variational Graph Auto-Encoder (DVGA) and Disentangled Graph Auto-Encoder (DGA). 
Initially, we design a Disentangled Graph Convolutional Network (DGCN) employing dynamic assignment mechanism \cite{Zheng2021adversarial}, incorporating multiple channels to aggregate information related to each disentangled latent factor. 
Then, a component-wise flow is applied to each channel to enhance the expressiveness.
Furthermore, we design a factor-wise decoder, ensuring that if any factor contributes to a link between nodes, then those two nodes will be connected in the final prediction result. 
To encourage the independence of representations corresponding to different latent factors, we impose independence constraints on the mapping channels of different latent factors. Finally, the auto-encoder and independent regularization will be co-optimized in a unified framework, allowing the disentangled graph encoder to learn better disentangled graph representations and improve the overall performance of the model.
Our models differ from existing methods by learning disentangled representations from the graph. This enables us to explore the significance of each channel, leading to enhanced interpretability and the generation of persuasive and comprehensible link prediction results.

Our contributions can be summarized as follows:
\begin{enumerate}[\IEEEsetlabelwidth{3)}]
    \item We present two novel disentangled graph auto-encoder models: the Disentangled Variational Graph Auto-Encoder (DVGA) and the Disentangled Graph Auto-Encoder (DGA). The DVGA model incorporates a disentangled encoder, component-wise flow, factor-wise decoder, and an independence regularization term in the loss function to learn disentangled representations. Similarly, the DGA model is also introduced. As far as we know, we are one of the only researchers studying disentangled graph auto-encoder models.
    \item Our proposed disentangled graph convolutional network in the encoder captures multiple factors by learning disentangled latent factors on graphs, enabling multi-relation disentangling. We enhance the expressive capability of each channel representation by incorporating flow, hence enhancing the overall capability of the representation. The designed decoder is tailored for disentangled representations, improving its ability to predict graph structures.
    \item To validate the effectiveness of our proposed models, we conduct comprehensive experiments on both synthetic datasets and real-world datasets. The results on several datasets demonstrate that DVGA and DGA achieves state-of-the-art performance, significantly outperforming the baselines both quantitatively and qualitatively.
\end{enumerate}

\section{Related Work}
In this section, we give a brief review of the research topics that are relevant to our proposed method, including graph embedding, extended (variational) graph auto-encoder, and disentangled representation learning.
\subsection{Graph Embedding}
The goal of graph embedding is to embed graph-structured data into a low-dimensional feature space while preserving the graph information. We classify graph embedding algorithms into two main types: topological embedding methods and content-enhanced embedding methods \cite{Cai2018a,Pan2018adversarially}.

Topological embedding methods only utilize the topological structure of the graph, aiming to maximally preserve the topological information. The DeepWalk model, introduced by \cite{Perozzi2014deepwalk}, learned node embeddings by random walks. Later, various probabilistic models have been proposed, such as LINE \cite{Tang2015line} and node2vec \cite{Grover2016node2vec}. Matrix factorization methods like HOPE \cite{Ou2016asymmetric} and M-NMF \cite{Wang2017community} have been developed to learn the latent representation of the graph, which leveraged the adjacency matrix. 
Deep learning models \cite{Zaiqiao2019coembedding} have also been employed for graph embedding, such as SDNE \cite{Wang2016structural} and DNGR \cite{Cao2016deep}.

Content-enhanced embedding methods
leverage both topological information and content features. TADW \cite{Yang2015network} extended the matrix factorization model to support embedding of attribute information. UPP-SNE \cite{Zhang2017user} employed an approximate kernel mapping method to improve the learning of user embeddings in social networks using user profile features. Deep learning models such as GCN \cite{Kipf2016semi}, GraphSAGE \cite{Hamilton2017inductive}, GAT \cite{Velickovic2017graph} and GSN \cite{Bouritsas2022improving} learned node representations through supervised tasks. Some works \cite{Kipf2016variational,Simonovsky2018graphvae} were based on generative models, specifically using Variational Auto-Encoder (VAE) \cite{Kingma2013auto} on graphs for learning representations. 
A comprehensive discussion on this topic will be provided in the subsequent subsection.
\subsection{Extended (Variational) Graph Auto-Encoder}
With the rapid development of deep learning, there has been an increasing focus on research in graph representation learning using generative models. Inspired by AE and VAE \cite{Kingma2013auto},  
Variational Graph Auto-Encoder (VGAE) and the Graph Auto-Encoder (GAE) \cite{Kipf2016variational} utilized GCN as encoders and inner products as decoders. They accomplished link prediction tasks while learning embeddings for each node in the graph.

In recent years, significant progress has been made in enhancing the structure of these two models. LGAE \cite{Salha2021simple} employed significantly simpler and more interpretable linear models as encoder. NF-VGA \cite{Shan2020nf} designed a prior-generative module to generate a flexible distribution as the prior for latent representations. SIG-VAE \cite{Hasanzadeh2019Semi} enhanced the generative modeling by employing a hierarchical variational framework.
In terms of decoder, TVGA \cite{Shi2020effective} leveraged the triadic closure property to introduce a triad decoder.
Graphite \cite{Grover2019graphite} employed an innovative iterative refinement strategy for graphs, inspired by low-rank approximations. In terms of model training improvements, ARGA and ARVGA \cite{Pan2018adversarially} introduced a novel adversarial graph embedding method.

There have also been other breakthroughs in terms of refining priors and other components of the models \cite{Davidson2018hyperspherical,Sun2021interpretable,Khan2021epitomic}. However, current approaches based on VGAE or GAE have only explored general settings in handling entanglement situations. Their encoders, lacking the integration of disentangled representation learning module, are challenged in recognizing and disentangling the heterogeneous latent factors hidden in the observed graph-structured data. Thus, these holistic encoding approaches exhibit limited capability in learning graph representations and often lead to less satisfying performance in link prediction tasks and other downstream applications.
\subsection{Disentangled representation learning}
The goal of learning disentangled representation is to obtain a factorized representation that can effectively identify and disentangle the underlying factors in observed data \cite{Bengio2013}. Disentangled representation learning has become a prominent challenge in machine learning due to its ability to generate robust and interpretable representations. 
The majority of existing research can be found in the field of computer vision, where variational methods based on VAE are widely employed for disentangled representation learning in images \cite{Higgins2016beta,Trauble2021on,Locatello2020weakly,Shen2022weakly}. This is mainly achieved by imposing independent constraints on the posterior distribution of the latent variable through KL divergence. 

There has been a recent surge of interest in studying disentangled representation learning for graph-structured data. The connections between nodes in a homogeneous graph remain heterogeneous and entangled, but they are only represented as single binary-valued edges. However, the edges in the graph often contain substantial relationship information that goes beyond simple binary indicators of structural connectivity. This motivation prompts us to implicitly uncover the underlying relationships between entities. DisenGCN \cite{Ma2019disentangled} is a notable and original advancement in the field of graph disentanglement. 
However, it did not take into account multiple potential relationships between entities. Later, ADGCN \cite{Zheng2021adversarial} considered the existence of multiple relationships between nodes and utilized an adversarial regularizer to enhance separability among different latent factors.
Related works mainly focused on node-level \cite{Guo2022learning,Liu2020independence}, edge-level \cite{Zhao2022exploring,Wu2021learning}, and graph-level \cite{Wu2022multi,Yang2020factorizable} disentanglement representation learning. Furthermore, another line of work employed a self-supervised learning strategy to learn disentangled representations \cite{Xiao2022decoupled,Li2021disentangled,Zhang2023contrastive}. 
Simultaneously, there has been a significant amount of study on disentangled representation learning for heterogeneous graphs \cite{Wu2021disenkgat,Geng2022disentangled}. Nevertheless, these works either typically rely on supervised labels and largely focus on node classification tasks, or they concentrate on other specific application domains. In contrast, we emphasize the use of generative models to learn disentangled node embedding for graph-structured data and focus on link
prediction and other downstream tasks.

Currently, there is limited research on the study of disentangled representation learning for graphs using generative models, specifically (variational) graph auto-encoder. There is just one study that is almost contemporary but specifically deals with the task of link prediction, as mentioned in \cite{Fu2023variational}. The authors employed a neighborhood routing method \cite{Ma2019disentangled} in their suggested model to obtain disentangled node embeddings. However, a drawback of this approach is its assumption that interactions between nodes are solely influenced by a single factor. In contrast, our models not only utilize a more general mechanism for learning representations but also boost the inference model's capacity through flow models. In addition, we construct a factor-wise decoder that is more closely aligned with the link prediction task that is stressed in generative models. The aim of our study is to learn disentangled node representations. Not only does this enhance performance in link prediction, but it also demonstrates excellent results in other subsequent downstream tasks.
\section{Preliminaries and Problem Formulation}
To better describe the issues that our paper focuses on and present our work, we first briefly review the concepts of GAE and VGAE in Section \ref{preliminaries} and then formulate the problems in Section \ref{Problem Formulation}.
\subsection{Preliminaries on (Vriational) Graph 
Auto-Encoder}\label{preliminaries}
VGAE and GAE, introduced by \cite{Kipf2016variational}, are specifically designed for unsupervised learning on graph-structured data. A remarkable characteristic is their natural ability to incorporate node features, resulting in a significant enhancement in predictive performance on diverse tasks across many benchmark datasets.

We will primarily focus on undirected graphs, though the approach can be easily extended to directed graphs. Let us begin by introducing some notations. Consider a graph $G = (V, E, X)$, where $V$ is the set of $N$ nodes, and $E$ is the set of edges. For any distinct nodes $i$ and $j$, if there's an edge connecting them, we denote it as $(i, j) \in E$. The topological structure of the input graph $G$ can be expressed using an adjacency matrix $A \in \mathbb{R}^{N\times N}$ where $A_{i,j}=1$ if $(i, j) \in E$, and $A_{i,j}=0$ otherwise. Its degree matrix is $D$. $X\in \mathbb{R}^{N \times f}$ is the node feature matrix, where each node $i \in V$ has a feature vector $x_i \in \mathbb{R}^f$, and $f$ is the dimension of raw features per node. The matrix for graph embeddings is $Z \in \mathbb{R}^{N\times d}$, with $z_i \in \mathbb{R}^d$ representing the embedding of node $i$.

GCN is used as the encoder in VGAE and GAE, which is designed specifically for processing and extracting information from graph-structured data. For a $L$-layer GCN, its message propagation rule is defined by the following formula:
\begin{equation}
 H^{(l+1)}=f_\theta(H^{(l)},A)={\rm Relu}(\tilde{D}^{\frac{1}{2}}\tilde{A}\tilde{D}^{-\frac{1}{2}}H^{(l)}W^{(l)}).
\end{equation}
Here, $\theta$ is the parameters of the encoder, $\tilde{A}=A+I_N$, $I_N$ is the identity matrix, $\tilde{D}_{ii}=\sum_{j=1}^{N}\tilde{A}_{ij}$ and $W^{(l)}$ is the parameters of weight matrix. ${\rm Relu}(\cdot)$ denotes the Relu activation function. $H^{(l)}$ is the matrix of activations in the $l$-th layer, $H^{(0)}=X$ and $H^{(L)}=Z$.
In GAE, \cite{Kipf2016variational} used two-layer GCN as the encoder:
\begin{equation}
H^{(1)}=f_\theta(X,A), Z=f_\theta(H^{(1)},A).
\label{GAE_encoder}
\end{equation}
In VGAE, the posterior distribution of the embedding vectors is given by the following formulation:
\begin{gather}
 q_\theta(Z|X,A)= \prod_{i=1}^Nq_\theta(z_i|X,A), \label{VGAE_encoder1}\\
 {\rm{with}}\quad q_\theta(z_i|X,A)=\mathcal{N}(\mu_i,{\rm{diag}}(\sigma_i^2)),\
 \label{VGAE_encoder2}
\end{gather}
where $\mu_i$ and ${\rm log}\,\sigma_i^2$ are learned from two GCNs with shared first-layer training parameters. Samples of $q(Z|X,A)$ can be obtained by using the reparameterization trick \cite{Doersch2016tutorial}.

The decoder relies on the inner product of the latent representations of each pair of nodes. In the case of GAE, the reconstruction of the adjacency matrix ${A}$ is formulated as follows:
\begin{equation}
 {A}=\sigma(ZZ^T).
\end{equation}
where $\sigma$ is the sigmoid function. The decoder for VGAE is probabilistic and can be expressed as follows:
\begin{gather}
 p({A}|Z)=\prod_{i=1}^N\prod_{i=1}^Np({A}_{ij}|z_i,z_j),\label{Eq: reconstruct}\\
{\rm{with}} \quad p({A}_{ij}|z_i,z_j)=\sigma(z_i^Tz_j).
\end{gather}

The training objective of VGAE is to recover information about the graph from the embedding matrix $Z$. Therefore, our objective is to estimate the model parameters by maximizing the logarithm of the marginal distribution, which can be lower-bounded by ELBO:
\begin{align}
     \log p_\theta(A|X) =& \log \int q_\theta(Z|X,A) \frac{p(Z)p(A|Z)}{q_\theta(Z|X,A)}dZ \\
    \geq& \int q_\theta(Z|X,A) \log \frac{p(Z)p(A|Z)}{q_\theta(Z|X,A)}dZ \\
    =& \mathbb{E}_{Z \sim q_\theta(Z|X,A)} \log p(A|Z) \nonumber\\
    &- D_{KL}(q_\theta(Z|X,A)|| p(Z)).\label{Eq:ELBO}
\end{align}
Here, the inequality is derived from Jensen's inequality \cite{Weisstein2006jensen}, and $D_{KL}(\cdot||\cdot)$ represents the Kullback-Leibler (KL) divergence \cite{Kullback1951oninformation}. We set $p(Z)$ to be a Gaussian normal distribution, i.e., 
\begin{equation}
    p(Z) = \prod_{i=1}^N p(z_i) = \prod_{i=1}^N \mathcal{N}(z_i | 0, \rm{I}).
\label{Eq:prior}
\end{equation}
For GAE, the non-probabilistic variant of VGAE, the training objective involves a simple reconstruction loss:
\begin{equation}
 \mathcal{L}_{recon}=-\mathbb{E}_{Z \sim q_\theta(Z|X,A)} \log p(A|Z).\label{Eq:recon loss}
\end{equation}

\subsection{Problem Formulation}\label{Problem Formulation}
For (variational) graph auto-encoder, the objective is to learn a graph encoder $f_\theta(\cdot)$ with parameters $\theta$ that outputs interpretable low-dimensional stochastic latent variables $Z = f_\theta(G) \in \mathbb{R}^{N\times d}$ ($d \ll n$). Simultaneously, such learned node representations are capable of tackling a wide variety of tasks related to nodes. 

In our study, we aim to improve the link prediction performance in (variational) graph auto-encoder by improving its overall framework, aligning with the experiments conducted in \cite{Kipf2016variational}. To achieve this, we will learn disentangled node representations using the disentangled graph encoder $f_\theta$. 

For each node $u$, the output $z_u$ from the disentangled graph encoder will be a disentangled representation. Specifically, assuming there are $K$ latent factors to be disentangled, $z_u$ is composed of $K$ independent components, i.e., $z_u=\left[z_{u,1},z_{u,2},\dots,z_{u,K}\right]$, where $z_{i,k}\in \mathbb{R}^{\frac{d}{K}}$ ($ 1\leq k \leq K$). The $k$-th component $z_{u,k}$ is used to characterize aspects of node $u$ that are related to the $k$-the latent factor. In this context, we assume the underlying factors generating observed graph-structured data are mutually independent which is well accepted in the field of disentangled representation learning \cite{higgins2018towards,Bengio2013}. Our goal is to learn factorized and expressive representations where each component is independent and exclusively corresponds to a single ground-truth factor behind the graph formulation. These representations will enhance performance in widely-used graph analytic tasks.

\begin{figure*}[!t]
\begin{center}
\includegraphics[width=\textwidth]{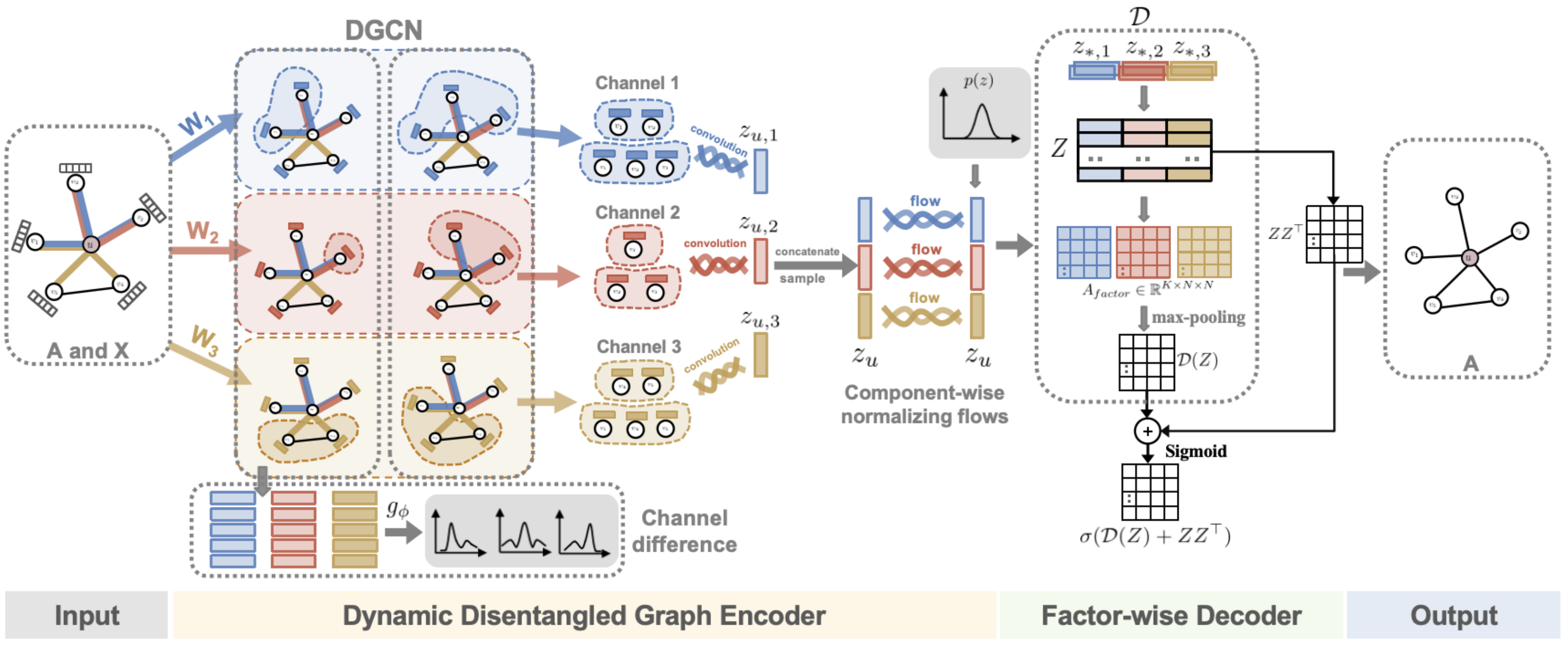}
\end{center}
\caption{
The whole framework of the proposed DVGA. It takes in nodes and their neighbors along with their feature vectors. The dynamic disentangled encoder utilizes DGCN, which employs a dynamic assignment mechanism that considers neighborhoods induced by $K$ distinct factors and neighborhoods in the $k$-th component space. After channel-wise convolution, the resulting component representations are merged and fed into flows for learning expressive disentangled node representations, which are then sent into a factor-wise decoder to reconstruct the adjacency matrix. A classifier $g_\phi$ is employed to promote the independence of component representations among several latent factors. The joint optimization of the factor-wise auto-encoder and independence regularization enhances the disentanglement. In this example, we assume the existence of three latent factors, each of which corresponds to one of the three channels. }
\label{Fig:model}
\end{figure*}
\section{Proposed Mechanism}
This section presents the DVGA model, with its framework illustrated by Figure \ref{Fig:model}. In the end, we also introduce its non-probabilistic variant, the DGA model.

\subsection{Dynamic Disentangled Graph Encoder}\label{Section:encoder}
Here, we introduce the disentangled graph encoder that can learn disentangled node embeddings. It will consist of a disentangled graph convolutional neural network, which we denote as DGCN. We assume that each node consists of $K$ independent components, indicating the presence of K latent factors that need to be disentangled. For a node $u\in V$, its hidden representation $z_u$ is expressed as $\left[z_{u,1},z_{u,2},\dots,z_{u,K}\right]$, where $z_{u,k}\in \mathbb{R}^{\frac{d}{K}}$ ($ 1\leq k \leq K$) is used to describe $k$-th aspect of node $u$. 

Firstly, we project the feature vector $x$ into $K$ different subspaces to initialize the embedding vectors for $K$ channels:
\begin{equation}
 c_{u,k}=\frac{W_k^{\top}x_u+b_k}{\lVert W_k^{\top}x_u+b_k \rVert_2},
 \label{Eq:K subspaces}
\end{equation}
where $W_k \in \mathbb{R}^{n\times \frac{d}{K}}$ and $b_k \in \mathbb{R}^{\frac{d}{K}}$ are the parameters of $k$-th subspace.
For a comprehensive understanding of aspect $k$ for node $u$, we need to extract information from its neighborhood.
This involves constructing  $z_{u,k}$ using both $c_{u,k}$ and $\{c_{v,k}: (u, v) \in E\}$.
So we provide three hypotheses for the relationships between nodes:

\noindent\textbf{Hypothesis 1:} The generation of links between nodes results from the interaction of various hidden factors.

\noindent\textbf{Hypothesis 2:} If there is similarity in the $k$-th component space between node $u$ and its neighboring node $v$, it indicates that the link between these two nodes may be attributed to factor $k$.

\noindent\textbf{Hypothesis 3:} If neighbors exhibit similarity in aspect $k$, forming a cluster in the $k$-th subspace, it is plausible that factor $k$ is responsible for connecting node $u$ with this specific group of neighbors.

\textbf{Hypothesis 1} aligns with our realistic understanding of the reasons behind the formation of graphs. It suggests that the links between nodes can be influenced by various factors. In comparison to previous research on disentanglement, this hypothesis is closer to real-world situations \cite{Liu2020independence,Ma2019disentangled}. For example, in a friendship network, the acquaintance between two individuals may result from relationships such as colleagues, classmates, or shared interests. \textbf{Hypotheses 2} and \textbf{3} correspond to the first-order and second-order proximities within the topological structure influenced by factor $k$, respectively \cite{Tang2015line}. This insight inspires us to explore $K$ subspaces to identify latent factors.

Building upon these three hypotheses, we can learn the disentangled representations of nodes by leveraging dynamic assignment mechanism \cite{Zheng2021adversarial}, which is a disentangle layer:
\begin{gather}   
    z_{u,k}^{t} = c_{u,k}+\alpha\sum_{v:(u,v)\in E}p_{u,v}^{k,t-1}c_{v,k}+\beta\sum_{v:(u,v)\in E}q_{v,u}^{k,t-1}c_{v,k}^t\label{Eq:z_iteration}\\
    p_{u,v}^{k,t} = \frac{{\rm exp}(c_{v,k}^{\top}z_{u,k}^t)}{\sum_{k=1}^K {\rm exp}(c_{v,k}^{\top}z_{u,k}^t)}\label{Eq:p}\\  
    q_{v,u}^{k,t} = \frac{{\rm exp}(c_{v,k}^{\top}z_{u,k}^t)}{\sum_{v:(u,v)\in E} {\rm exp}(c_{v,k}^{\top}z_{u,k}^t)}\label{Eq:q}\\
    z_{u,k}^{t}=\frac{z_{u,k}^{t}}{\lVert z_{u,k}^{t} \rVert_2}\label{Eq:embedding_normalize}
\end{gather}
where iteration $t=1,2,\dots,T$. $p_{u,v}^{k}$ represents the probability of the connection between nodes $v$ and $u$ attributed to the $k$-th factor, while $q_{v,u}^{k}$ denotes the importance of aggregating node $v$ with respect to node $u$ in the $k$-th component space. They satisfies $p_{u,v}^{k} \geq 0$, $q_{v,u}^{k} \geq 0$, $\sum_{k=1}^{K}p_{u,v}^{k}=1$ and $\sum_{k=1}^{K}q_{v,u}^{k}=1$. $\alpha$ and $\beta$ serve as trade-off coefficients and we treat them as trainable parameters and adjust their range to be within $(0,1)$. The dynamic assignment mechanism will iteratively infer $p_{u,v}^{k}$ ,$q_{v,u}^{k}$ and construct $z_{u,k}$. Meanwhile, this mechanism enables the coexistence of multiple relationships between nodes. It's important to note that there are a total of $L$ disentangle layers, and in each layer $l \leq L-1$, the value of $c_{u,k}$ is ultimately assigned as $z_{u,k}^{T}$. 

The final disentangled embedding of node $u$ output by DGCN is the concatenation of components learned across all $K$ channels, i.e., 
\begin{equation}
z_u={\rm concatenate}(z_{u,1},z_{u,2},\dots,z_{u,K}). 
\end{equation}
By applying Eqs. (\ref{GAE_encoder}), (\ref{VGAE_encoder1}), (\ref{VGAE_encoder2}) and replacing GCN with DGCN, we finally obtain the dynamic disentangled graph encoder. 

In contrast to existing holistic graph encoders, this dynamic disentangled graph encoder comprises $K$ channels, enabling the identification of complex heterogeneous latent factors and the depiction of multiple relation aspects.
\subsection{Enhancing Posterior Flexibility through Component-wise Normalizing Flows}
It is evident that when the inference process depends exclusively on the dynamic disentangled graph encoder, DVGA will generally provide a distribution that is factorized. This suggests that the individual component representation $z_{u,k}$ is not only independent of each other but also mutually independent for each $d$-dimensional vector. It is unnecessary and undesirable to have the latter because we aim for each $d$-dimensional component to effectively capture sufficient information about each factor. 
In this section, we aim to improve the flexibility of each $d$-dimensional component representation so that it can effectively capture the complex properties of the factors.

Normalizing Flows (NF) \cite{Papamakarios2017masked} are commonly employed to expand the range of posterior distribution families. They impose explicit density functions for the mixing distributions in the hierarchy\cite{Hasanzadeh2019Semi,Shan2020nf}. 
To better learn the representation of each factor, we introduce a flow-based, component-wise distribution as the posterior distribution for the latent variables in DVGA. 

The main flow process involves applying separate flows to the $K$ components of $z_u$, with each flow consisting of $M$ steps (e.g., 
$M$ = 4). Each step includes an affine coupling layer, which guarantees that the latent variable $z_{u,k}$ follows a flexible distribution. 
More precisely, we follow the following iteration process to apply flows to each component $1\leq k \leq K$, denoted as $m = 1,2,\dots,M$:
\begin{gather}
    z_{u,k,1:d'}^{(m)}=z_{u,k,1:d'}^{(m-1)},\label{Eq:flow1}\\
    z_{u,k,d'+1:d}^{(m)}=z_{u,k,d'+1:d}^{(m-1)}\odot {\rm exp}(s_k(z_{u,k,1:d'}^{(m-1)}))+t_k(z_{u,k,1:d'}^{(m-1)})\label{Eq:flow2}.
\end{gather}
Here, $s_k(\cdot)$ and $t_k(\cdot)$ denote transformation functions (mapping the input from $d'$ to $d-d'$ dimension space), $s(z_{u,k,1:d'}^{(m-1)})=\left[s_{1}(z_{u,k,1:d'}^{(m-1)}),s_{2}(z_{u,k,1:d'}^{(m-1)}),\dots,s_{d-d'}(z_{u,k,1:d'}^{(m-1)})\right]^\top$, $t(z_{u,k,1:d'}^{(m-1)})=\left[t_{1}(z_{u,k,1:d'}^{(m-1)}),t_{2}(z_{u,k,1:d'}^{(m-1)}),\dots,t_{d-d'}(z_{u,k,1:d'}^{(m-1)})\right]^\top$. The $\odot$ represents the Hadamard product. The output of the affine coupling layer is obtained by concatenating $z_{u,k,1:d'}^{(m)}$ and $z_{u,k,d'+1:d}^{(m)}$. We swap or randomly shuffle the dimensions to enhance the mixing of information between two affine coupling layers. 

This process transforms the posterior distribution $q(z|X, A)$ into a more complex distribution, allowing each $z_{u,k}$ to have sufficient expressive capacity while preserving the separability of information between each channel.

\subsection{Factor-wise Decoder}
Unlike the traditional decoder design found in the (variational) graph auto-encoder and other approaches that broadly utilize graph information to design decoder \cite{Hasanzadeh2019Semi,Grover2019graphite,Shi2020effective}, both DGA and DVGA utilize a novel factor-wise decoder to predict edge connections between nodes. This architecture promotes link prediction performance by encouraging the decoder to integrate information at the factor level.

As we already know, the formation of real-world graphs is typically influenced by multiple latent heterogeneous factors. Therefore, if any factor contributes to the presence of a connection between two nodes, an edge will be established between them in the final outcome. Given this insight, we make alterations to the conventional decoder. The structure of the proposed decoder is also illustrated in Figure \ref{Fig:model}. First, we calculate the cosine similarity between each pair of nodes $z_u=\left[z_{u,1},...,z_{u,K}\right]^{\top}$ and $z_v=\left[z_{v,1},...,z_{v,K}\right]^{\top}$ for each of the $K$ components. The output vector $A_{factor}$ is a three-dimensional matrix in $\mathbb{R}^{K\times N\times N}$. It represents the predicted connection status of nodes in $K$ component spaces that are related with $K$ factors. 
Each component space link yields a $N \times N$ matrix as the outcome. This vector $A_{factor}$ is then fused by performing a max-pooling operation to the prediction of each two-node connection of the $K$ channels. This process results in the final state of the edge connection. The entire block, denoted as $\mathcal{D}$, is ultimately merged with inner products constructed from $Z$. The additional inner product link operates similarly to the residual connections utilized in residual networks \cite{He2016deep}.
Finally, the output of the factor-wise decoder is as follows:
\begin{equation}
p({A}|Z)=\sigma(\mathcal{D}(Z)+ZZ^{\top}).
\label{Eq:factor-wise decoder}
\end{equation} 
This decoder function is an expansion of the inner product decoder, as it turns into the conventional inner product decoder when $\mathcal{D}$ outputs a zero mapping.

\subsection{ Evidence Lower Bound (ELBO)}
Now we describe the objective of DVGA.
The model parameters, denoted as $\theta$, are obtained by maximizing the lower bound of log-likelihood, as specified in Eq.(\ref{Eq:ELBO}): 
\begin{align}
\theta^{*}=&\mathop{\rm argmax}\limits_{\theta}\,\mathcal{L}_{ELBO}(\theta)\label{Eq: argmax1}\\
=&\mathop{\rm argmax}\limits_{\theta}\left[\mathbb{E}_{Z \sim q_\theta(Z|X,A)} \log\,p(A|Z) \right.\nonumber\\
&\left.- D_{KL}(q_\theta(Z|X,A)|| p(Z))\right]\label{Eq: argmax2}
\end{align}
The first term corresponds to the reconstruction loss for link prediction, as defined by Eq. (\ref{Eq: reconstruct}) and Eq. (\ref{Eq:factor-wise decoder}). The second term represents the KL divergence, where the prior $p(Z)$ is formulated by Eq. (\ref{Eq:prior}).

\noindent\textbf{Claim.} \textit{The} $q_\theta(Z|X,A)$ in Eq. (\ref{VGAE_encoder1}) \textit{ can be computed as follows:}
\begin{align}
\log\,q_\theta(z_u|X,A)=&
\log\,q_\theta(z^{(0)}_u|X,A)\nonumber\\
&-\sum_{m=1}^{M}\sum_{k=1}^K\sum_{i=d'+1}^d s_i(z_{u,k,1:d'}^{(m-1)})\label{Eq:log_q 2}
\end{align}
\textit{where} $u\in V$, $z_u=z_u^{(M)}$ \textit{and} $z_u^{(0)}$ \textit{is the latent variable output from the dynamic graph encoder. }$f_{k}^{(m)}$ \textit{represents the complete function of the flow for channel} $k$ \textit{from step} $m-1$ \textit{to} $m$ \textit{as described in }Eqs. (\ref{Eq:flow1}) \textit{and} (\ref{Eq:flow2}).
\begin{proof} Eqs. (\ref{Eq:flow1}) and (\ref{Eq:flow2}) provide the functions of the flow from step $m-1$ to $m$ for the component $z_{u,k}$, denoted as $f_{k}^{(m)}$. Set $f^{(m)}=\left[f_{1}^{(m)},f_{2}^{(m)},\dots,f_{K}^{(m)}\right]$. Considering the following equation,:
\begin{eqnarray}
z^{(m)}_u&=&\left[z_{u,1}^{(m)},z_{u,2}^{(m)},\dots,z_{u,K}^{(m)}\right]\nonumber\\
&=&\left[f_{1}^{(m)}(z_{u,1}^{(m-1)}),f_{2}^{(m)}(z_{u,2}^{(m-1)}),\dots,f_{K}^{(m)}(z_{u,K}^{(m-1)})\right]\nonumber\\
&=&f^{(m)}(z_u^{(m-1)}),
\end{eqnarray}
then, we can compute the following Jacobian matrix:
\begin{equation}
{\rm det}(J_{f^{(m)}}(z_{u,1}^{(m-1)},z_{u,2}^{(m-1)},\dots,z_{u,K}^{(m-1))})
=\prod_{k=1}^{K}\diff{f_{k}^{(m)}}{z_{u,k}^{(m-1)}} \label{Eq:jacobian_prod}
\end{equation}
Note that in Eq. (\ref{Eq:jacobian_prod}), we can further use the fact that $\diff{f_{k}^{(m)}}{z_{u,k}^{(m)}}$ is a lower triangular matrix, yielding:
\begin{equation}
\diff{f_{k}^{(m)}}{z_{u,k}^{(m-1)}}=\prod_{i=d'+1}^d {\rm exp}(s_i(z_{u,k,1:d'}^{(m-1)}))
\end{equation}
Finally, we can compute the ultimate formulation as follows:
\begin{align}
\log\,q_\theta(z_u|X,A)=&\log\,(q_\theta(z_{u}^{(0)}|X,A)\left|{\rm det}\,\frac{\partial{z_{u}^{(0)}}}{\partial{z_u}}\right|)\\
=&\log\,q_\theta(z_{u}^{(0)}|X,A)\nonumber\\
&-\sum_{m=1}^{M}\sum_{k=1}^K\log\,\left|\frac{{\rm d}f_{k}^{(m)}}{{\rm d}z_{u,k}^{(m-1)}}\right|\\
=&\log\,q_\theta(z_{u}^{(0)}|X,A)\nonumber\\
&-\sum_{m=1}^{M}\sum_{k=1}^K\sum_{i=d'+1}^d s_i(z_{u,k,1:d'}^{(m-1)}).
\end{align}
Thus, the proof is complete.\end{proof}

In the above proof, the simplicity of the computations are attributed to the affine coupling flow used in Eqs. (\ref{Eq:flow1}) and (\ref{Eq:flow2}). In the end, after computing $\log\,q_\theta(z_u|X,A)$, we can use Eq. (\ref{VGAE_encoder1}) and employ the reparameterization trick \cite{Doersch2016tutorial} to get $D_{KL}(q_\theta(Z|X,A)|| p(Z))$, so Eq. (\ref{Eq: argmax2}) can be calculated.
\subsection{Statistical Independence of Mapping Subspaces}
Our main goal is to empower the graph encoder to generate disentangled representations $z_{u}=\left[z_{u,1},z_{u,2},\dots,z_{u,K}\right]$ for each latent factor. The $K$ different factors extracted by the dynamic routing mechanism are intended to focus on different connecting causes, with each of the $K$ channels capturing mutually exclusive information related to every latent factors. This highlights the significance of encouraging statistical independence among the disentangled representations in order to improve the process of disentanglement. Therefore, it is crucial to ensure that the focused perspectives of the $K$ subspaces in Eq. (\ref{Eq:K subspaces}) are distinct before performing dynamic assignment mechanism. Hence, we propose imposing independent constraints on these $K$ subspaces.

Unfortunately, it is not trivial to obtain the solution that all disentangled subspaces are maximally different from each other. Therefore, we propose imposing independence constraints on the initial embedding vectors obtained from projections in $K$ different subspaces, as expressed in Eq. (\ref{Eq:K subspaces}). We estimate the solution by giving distinct labels to embedding vectors belonging to the same subspace and maximizing the mapping subspaces as a problem of vector classification. The discriminator, as denoted by the Eq. (\ref{Eq:dicriminator}), is used to differentiate the label associated with a particular embedding vector:
\begin{equation}
    D_k = {\rm Softmax}\left(g_\phi(c_{u,k})\right) , k=1,2,\dots,K, \forall u\in V.
\label{Eq:dicriminator}
\end{equation}
This discriminator, represented as $g_\phi$, comprises a fully connected neural network, and its parameters are denoted as $\phi$. It takes the embedding vector $c_{u,k}$ as inputs and generates a subspace-specific label $D_k$.

We formulate the following loss function to train the discriminator:
\begin{equation}
    \mathcal{L}_{reg}(\phi)= -\frac{1}{N}\sum_{i=1}^{N}\left(\sum_{e=1}^{N_k}\mathbbm{1}_{k=e}\log\,(D_k^i\left[e\right])\right)\label{Eq:regularizer}
\end{equation}
where $N$ represents the number of nodes, $N_k=K$ is the count of latent factors, $D_k^i$ is the distribution of nodes $i$ and $D_k^i\left[e\right]$ denotes the probability that the generated embedding vector has label $e$. The indicator function $\mathbbm{1}_{k=e}$ takes the value of one when the predicted label is correct.

Explicitly strengthening the independence of the disentangled node representations will improve the graph encoder's ability to capture different information across several latent factors, potentially boosting performance in subsequent tasks.  
\begin{algorithm}[!t]
    \caption{The Training Procedure of DVGA}\label{alg:DVGA}
    \noindent\textbf{Input:} Graph $G=(V,E,X)$, the number of disentangle layers $L$, the number of iterations $T$, the number of steps for flows $M$, affine coupling parameters of flows $d'$, convolution coefficients $\alpha$ and $\beta$, the number of channel $K$, latent dimension $d$, regularization parameter $\lambda$ and training epochs $E$.  \\
    \textbf{Parameters:} \\
    Weight matrix $W_k\in \mathbb{R}^{n\times\frac{d}{K}}$ for $k=1,2,\dots,K$;\\
    Bias $b_k\in \mathbb{R}^{\frac{d}{K}}$ for $k=1,2,\dots,K$;\\
    Scale parameters of neural network for flow $s_k:\mathbb{R}^{d'}\rightarrow \mathbb{R}^{d-d'}$ for $k=1,2,\dots,K$;\\
    location parameters of neural network for flow $t_k:\mathbb{R}^{d'}\rightarrow \mathbb{R}^{d-d'}$ for $k=1,2,\dots,K$;\\
     \Comment{All these parameters above are collectively denoted as $\theta$;}\\
    Parameters of a neural network $g_\phi:\mathbb{R}^{\frac{d}{K}}\rightarrow \mathbb{R}^{K}$. 
    \begin{algorithmic}[1]
    \Function{Encoder}{$G=(V,E,X)$,$A$}
    \State $\boldsymbol{\mu}$ = \Call{DGCN}{$G$,$A$},
        \State ${\rm log}\,\boldsymbol{\sigma}$ = \Call{DGCN}{$G$,$A$},
        \State $\boldsymbol{\epsilon}\sim \mathcal{N}(0,\rm{I})$.
        \State $Z = \boldsymbol{\mu}+\boldsymbol{\epsilon}\circ \boldsymbol{\sigma}$. \Comment{$Z$ is assigned the value of $\boldsymbol{\mu}$ at test time. The symbol $\circ$ denotes element-wise multiplication.}   
        \State \textbf{return} $Z$
    \EndFunction
    \Function{Component-wise Flow}{$z_{u,k}$}
        \For{$m=1,2,\dots,M$}
            \State Update $z_{u,k}$ by Eqs. (\ref{Eq:flow1}) and (\ref{Eq:flow2}).
        \EndFor
        \State  \textbf{return} $z_{u,k}$
    \EndFunction
    \Function{Decoder}{$Z$}
        \State $Z_k\gets$ the submatrix corresponding to the $k$-th channel in $Z$.
        \State $A_{factor}^k=\cos(Z_k,Z_k)\in \mathbb{R}^{N\times N}$.
        \State $\mathcal{D}(Z)\gets maxpooling(A_{factor}^1,A_{factor}^2,\dots,A_{factor}^K)$.
        \State $\hat A = \sigma(\mathcal{D}(Z)+ZZ^{\top})$.
        \State \textbf{return} $\hat A$
    \EndFunction
    \For{epoch $=1$ to $E$}
          \State $Z, D_{KL}$ = \Call{Encoder}{$G$,$A$}.
          \State $Z_k\gets$ the submatrix corresponding to the $k$-th channel in $Z$.
          \State $Z_k$ = \Call{Component-wise Flow}{$\left\{z_{u,k}\right\}$}, $Z_k=\left\{z_{u,k}\right\}$.
          \State $Z$= Stack($\left[Z_1, Z_2, \dots, Z_K\right]$, dim=0).
          \State $p(A|Z) = \Call{Decoder}{Z}$.
          \State Calculate $D_{KL}(q_\theta(Z|X,A)||\mathcal{N}(\textbf{0},\textbf{I}))$ by Eq. (\ref{Eq:log_q 2}).
          \State Calculate $\mathcal{L}_{ELBO}(\theta)$ by Eq. (\ref{Eq:ELBO}) and the independence regularizer $\mathcal{L}_{reg}(\phi)$ by Eq. (\ref{Eq:regularizer}).
          \State Update $\theta$ and $\phi$ to minimize $\mathcal{L} = -\mathcal{L}_{ELBO}(\theta)+\lambda\mathcal{L}_{reg}(\phi)$ by Eq. (\ref{Eq:objective all}) .
   \EndFor
   \end{algorithmic}
\end{algorithm}
\subsection{Optimization}
Finally, our aim is to learn the parameters $\theta$ of the inference model and $\phi$ of the regularization term for DVGA in a unified framework. By employing gradient descent, we optimize the combined objective function that integrates the Evidence Lower Bound (ELBO) and the independence regularizer:
\begin{equation}
    \mathop{\rm min}\limits_{\theta,\phi}\,-\mathcal{L}_{ELBO}(\theta)+\lambda\mathcal{L}_{reg}(\phi),\label{Eq:objective all}
\end{equation}
where $\lambda$ is a hyper-parameter that controls the balance between $-\mathcal{L}_{ELBO}(\theta)$ and the impact of the regularizer. $\mathcal{L}_{reg}(\phi)$ is the loss of the discriminator, designed to encourage statistical independence among the disentangled node representations. The detailed training procedure for the DVGA model is shown in Algorithm \ref{alg:DVGA}.

For DGA, we employ the dynamic graph encoder in Section \ref{Section:encoder} to get latent representations, followed by the use of the factor-wise decoder in Eq. (\ref{Eq:factor-wise decoder}) for link prediction. The corresponding loss is given by:
\begin{equation}
    \mathop{\rm min}\limits_{\theta,\phi}\,\mathcal{L}_{recon}(\theta)+\lambda\mathcal{L}_{reg}(\phi),\label{Eq:DGA loss all}
\end{equation}
where the first term corresponds to the reconstruction loss Eq. (\ref{Eq:recon loss}) for link prediction, typically implemented as a binary cross-entropy loss. 
The training process for the DGA model is outlined in Algorithm \ref{alg:DGA}.
\begin{algorithm}[!t]
    \caption{The Training Procedure of DGA}\label{alg:DGA}
    \noindent\textbf{Input:} Graph $G=(V,E,X)$, the number of disentangle layers $L$, the number of iterations $T$, convolution coefficients $\alpha$ and $\beta$, the number of channel $K$, latent dimension $d$, regularization parameter $\lambda$ and training epochs $E$.  \\
    \textbf{Parameters:} \\
    Weight matrix $W_k\in \mathbb{R}^{n\times\frac{d}{K}}$ for $k=1,2,\dots,K$;\\
    Bias $b_k\in \mathbb{R}^{\frac{d}{K}}$ for $k=1,2,\dots,K$;\\
    \Comment{All these parameters above are collectively denoted as $\theta$;}\\
    Parameters of a neural network $g_\phi:\mathbb{R}^{\frac{d}{K}}\rightarrow \mathbb{R}^{K}$. 
    \begin{algorithmic}[1]
    \For{epoch $=1$ to $E$}
          \State $Z, D_{KL}$ = \Call{DGCN}{$G$,$A$}.
          \State $p(A|Z) = \Call{Decoder}{Z}$.
          \State Calculate $\mathcal{L}_{recon}(\theta)$ by Eq. (\ref{Eq:recon loss}) and the independence regularizer $\mathcal{L}_{reg}(\phi)$ by Eq. (\ref{Eq:regularizer}).
          \State Update $\theta$ and $\phi$ to minimize $\mathcal{L} = -\mathcal{L}_{recon}(\theta)+\lambda\mathcal{L}_{reg}(\phi)$ by Eq. (\ref{Eq:DGA loss all}).
   \EndFor
   \end{algorithmic}
\end{algorithm}
\section{Discussions}
\subsection{Time Complexity Analysis}
We have conducted a theoretical analysis of the time complexity of the proposed DVGA model, which is $O((K^2+fd+LT(C\frac{d}{K}+d))N+Md+(d+K)N^2) = O((K^2+(f+LT)d+LTC)N+(d+K)N^2+Md)= O(N^2)$. Here, $N$ denotes the number of nodes in the graph, $d$ is the dimensionality of latent variables, $f$ represents the dimensionality of node input features, $L$ is the number of layers in the disentangled layers employed by the dynamic disentangled graph encoder, $K$ indicates the number of latent factors and $T$ is the number of iterations for dynamic assignment. We define the maximum number of neighbors to be $C$. $d$, $T$, $L$, $C$ and $K$ are small constants. 

Specifically, the time complexity of our dynamic disentangled graph encoder is $O((fd+LT(c\frac{d}{K}+d))N)$. 
For the component-wise flow, $K$ channels are required to be implemented, each with a dimension of $\frac{d}{K}$, and $M$ steps, leading to a time complexity of $O(Md)$. The $O((d+K)N^2)$ is attributed to the factor-wise decoder, indicating that the computational cost is predominantly affected by the size of the graph-structured data, as demonstrated in the empirical study presented in Section \ref{Sec:Convergence Behavior and Complexity Analysis}. Nevertheless, we argue that our proposed DVGA remains reasonably efficient in practical applications, particularly when considering the substantial performance improvements validated in our experiments. The time complexity of computing the regularizer in our model is $O(NK^2)$. 

Similarly, we are able to get that the time complexity of DGA is $O((K^2+fd+LT(C\frac{d}{K}+d))N+(d+K)N^2) = O((K^2+(f+LT)d+LTC)N+(d+K)N^2)=O(N^2)$.
\subsection{Number of Parameters Analysis}
The number of parameters for the proposed DVGA is $O(fd+KMd'(d-d')) = O(fd+KMd^2)$, where 
$d'$ is employed in the affine coupling parameters of flows to determine the dimensions that remain unchanged from the previous step to the subsequent one. 

Specifically, the number of parameters for the dynamic disentangled encoder is $O(fd)$, the flow utilizes $O(KMd'(d-d'))$ parameters, and the independence regularization term involves $O(d)$ parameters. This parameter complexity is independent of the number of nodes and edges in the input graph, resulting in a relatively small value. 

Likewise, the number of parameters in DGA is $O(fd)$.
\section{Experiments}
In this section, we compare our method with other state-of-the-art methods  in three different applications, including unsupervised learning for link prediction and node clustering, as well as semi-supervised learning for node classification. We also qualitatively investigate our models' performance in factor disentanglement. 
\subsection{Link prediction}
Link prediction aims to predict the presence or absence of a connection between two nodes \cite{Loehlin2004latent}. It is one of the most common tasks in the representation learning of networks. By using the decoder in our model, we can effectively recover missing links or predict potential links through an analysis of the given graph.

\textit{1) Real-World Datasets:} We evaluate the effectiveness of our models  on three widely used benchmark citation network datasets: Cora, CiteSeer, and PubMed. In these datasets, nodes correspond to research papers, edges represent citations, and labels indicate the research areas \cite{Sen2008collective}. 
The features at the node level correspond to textual attributes found in the papers. Additionally, each node is assigned with a label indicating its class. We treat the graphs as undirected graphs. Table \ref{Tab:Statistics of citation data.} provides a summary of the dataset statistics.
\begin{table}[h]
\renewcommand{\arraystretch}{1}
\caption{Statistics for the citation network datasets.}
\begin{center}
\begin{tabular}{lccc}\hline
   Datasets &Cora &CiteSeer&PubMed \\\hline
   \#Nodes  &   2708   &    3327    &  19717 \\
   \#Edges  &   5429   &    4732    &  44338  \\
   \#Features  &   1433   &    3703    &  500  \\
   \#Classes &   7   &    6    &  3 \\\hline
\end{tabular}
\end{center}
\label{Tab:Statistics of citation data.}
\end{table}

\textit{2) Synthetic Dataset:} 
To explore the performance of our models on graphs with $m$ latent factors, we follow the methodology from \cite{Guo2022learning} for constructing synthetic graphs. Specifically, we create $m$ Erd\"{o}s-R\'{e}nyi random graphs, each consisting of 1000 nodes and 16 classes. Nodes within the same class are connected with a probability of $p$, while nodes from other classes are connected with a probability of $q$. Subsequently, by summing their adjacency matrices, we aggregate these generated graphs and set elements greater than zero to one. Then, we get the final synthetic graph with $m$ latent factors. This composite graph includes $16m$ classes and assigns $m$ labels to each node, which are obtained from the original labels of $m$ random graphs. Node representations are established based on the rows of the adjacency matrix. We fix the value of $q$ at $3e^{-5}$ and fine-tune $p$ to attain an average neighborhood size of around 40.

\textit{3) Baselines:} We evaluate the performance of DVGA and DGA in comparison to state-of-the-art baselines, including graph embedding methods, such as Spectral Clustering (SC) \cite{Tang2011leveraging}, DeepWalk \cite{Perozzi2014deepwalk}, and node2vec \cite{Grover2016node2vec}, as well as graph auto-encoder algorithms, including GAE/VGAE \cite{Kipf2016variational}, ARGA/ARVGA \cite{Pan2018adversarially}, LGAE/LVGAE \cite{Salha2021simple}, SIGVAE \cite{Hasanzadeh2019Semi} and GNAE/VGNAE \cite{Ahn2021variational}. Notably, SC, DeepWalk and node2vec lack the capability to use node features during the embedding learning process. Therefore, we evaluate them only on datasets without node features.

\textit{4) Evaluation metrics:} 
We employ two metrics, Area Under the ROC Curve (AUC) and Average Precision (AP), to evaluate the models' performance \cite{Kipf2016variational}. Each experiment is conducted five times, and the final scores are reported as mean values with standard errors. The real-world datasets are partitioned into training, testing, and validation sets. We augment the original graph with a balanced set of positive and negative (false) edges in the validation and testing sets. Specifically, the validation set comprises 5\% edges, the test set includes 10\% edges, and the remaining edges are used for training. Both the validation and testing sets include an equal number of non-edges. For synthetic datasets, we perform a random split, allocating 60\% to the training set, 20\% to the validation set, and another 20\% to the test set \cite{Guo2022learning}.

\textit{5) Parameter Settings:} We configure our DGA and DVGA models with \(d = K \Delta d\), where \(d\) represents the hidden dimension, \(K\) denotes the number of channels, and \(\Delta d = \frac{d}{K}\) is the output dimension of each channel. Hyper-parameter tuning is performed on the validation split of each dataset using Optuna \cite{Akiba2019optuna} for efficiency. Specifically, Optuna is executed for 100 trials for each configuration, and the hyper-parameter search space is defined as follows: dropout \(\sim \{0, 0.05, ..., 1\}\) with a step of 0.05, learning rate \(\sim [1e^{-3}, 1]\), the number of channels \(K \in \{2, 3, ..., 10\}\), the iteration of routing \(T \in \{1, 2, ..., 10\}\), the number of layers \(L \in \{1, 2, ..., 6\}\), and the regularization term \(\lambda \sim [1e^{-5}, 1]\). The output dimension of each channel is \(\Delta d = 16\). For DVGA, the number of flow steps \(\in \{1, 2, ..., 5\}\). The models are trained for 3000 epochs using the optimal hyper-parameters, and the performance is evaluated based on five test split runs. For the remaining baselines, we maintain consistency with our models in terms of the number of latent dimensions, while other parameters are set according to the configurations stated in the corresponding papers.

\begin{table*}[!tbp]
    \renewcommand{\arraystretch}{1}
    \caption{Results for Link Prediction of citation networks.($*$ denotes dataset without features, i.e., $X=\rm{I}$). Higher is better.}
    \centering
    \setlength\tabcolsep{1mm}{
    \scalebox{1}{
        \begin{tabular}{lcccccc}
        \hline
            \multirow{2}{*}{\textbf{Model}} & \multicolumn{2}{c}{\textbf{Cora*}} & \multicolumn{2}{c}{\textbf{CiteSeer*}} & \multicolumn{2}{c}{\textbf{PubMed*}}  \\ \cline{2-7}
             & Auc & AP & Auc & AP & Auc & AP   \\\hline
            SC  & $89.4 \pm 1.42$ & $89.8 \pm 1.44$ & $90.1 \pm 1.2$ & $89.2 \pm 1.31$ & $90.1 \pm 0.38$ & $87.5 \pm 0.65$  \\
            DW  & $85 \pm 1.29$ & $86.4 \pm 0.91$ & $89 \pm 0.99$ & $90.6 \pm 0.72$ & $91.7 \pm 0.11$ & $92.2 \pm 0.22$  \\
            node2vec & $85.7 \pm 1.09$ & $88.7 \pm 1.27$ & $90.3 \pm 1.22$ & $92.9 \pm 0.74$ & $93.9 \pm 0.13$ & $94.6 \pm 0.15$  \\\hline
            GAE & $83.8 \pm 1.37$ & $88.4 \pm 0.74$ & $76 \pm 1.11$ & $82.8 \pm 0.56$ & $81.2 \pm 0.33$ & $87.3 \pm 0.27$  \\
            LGAE & $85 \pm 1.21$ & $89.4 \pm 0.92$ & $79.5 \pm 1$ & $84.9 \pm 0.72$ & $83.5 \pm 0.45$ & $88.5 \pm 0.4$  \\
            ARGA & $76.4 \pm 2.62$ & $78 \pm 2.11$ & $71.5 \pm 0.87$ & $76.4 \pm 0.76$ & $79 \pm 0.4$ & $84.9 \pm 0.38$  \\
            GNAE & $87 \pm 1.26$ & $91.1 \pm 0.81$ & $79.2 \pm 1.2$ & $85.2 \pm 0.94$ & $84.4 \pm 0.13$ & $89.8 \pm 0.13$  \\
            DGA & $\boldsymbol{99.4 \pm 0.13}$ & $\boldsymbol{99.5 \pm 0.11}$ & $\boldsymbol{99.1 \pm 0.07}$ & $\boldsymbol{99.2 \pm 0.05}$ & $\boldsymbol{98.7 \pm 0.13}$ & $\boldsymbol{98.8 \pm 0.11}$  \\\hline
            VGAE & $84.8 \pm 1.26$ & $88.5 \pm 0.85$ & $76.4 \pm 0.92$ & $82.3 \pm 0.47$ & $82.3 \pm 0.43$ & $87.8 \pm 0.29$  \\
            LVGAE & $84.8 \pm 1.3$ & $88.8 \pm 0.98$ & $79.1 \pm 1.2$ & $84.2 \pm 0.91$ & $84.1 \pm 0.6$ & $88.9 \pm 0.42$  \\
            ARVGA & $68.9 \pm 4.07$ & $72.6 \pm 3.59$ & $71.4 \pm 1.74$ & $75.6 \pm 2.23$ & $84 \pm 0.57$ & $87.9 \pm 0.48$  \\
            SIGVAE & $85.1 \pm 0.8$ & $88.5 \pm 0.6$ & $77.6 \pm 1.58$ & $83.4 \pm 1.46$ & - & -  \\
            VGNAE & $86 \pm 0.97$ & $90.1 \pm 0.55$ & $78.7 \pm 1.46$ & $85 \pm 1.23$ & $84.6 \pm 0.44$ & $89.7 \pm 0.36$  \\
            DVGA  & $\boldsymbol{97 \pm 0.15}$ & $\boldsymbol{97.2 \pm 0.21}$ & $\boldsymbol{94.6 \pm 0.48}$ & $\boldsymbol{95.3 \pm 0.37}$ & $\boldsymbol{97.8 \pm 0.09}$ & $\boldsymbol{98 \pm 0.07}$  \\\hline
        \end{tabular}
    }}
    \label{Tab:link prediction citation network without feature}
\end{table*}
\begin{table*}[!tbp]
    \renewcommand{\arraystretch}{1}
    \caption{Results for Link Prediction of citation networks. Higher is better.}
    \centering
    \setlength\tabcolsep{1mm}{
    \scalebox{1}{
        \begin{tabular}{lcccccc}
        \hline
            \multirow{2}{*}{\textbf{Model}} & \multicolumn{2}{c}{\textbf{Cora}} & \multicolumn{2}{c}{\textbf{CiteSeer}} & \multicolumn{2}{c}{\textbf{PubMed}}  \\ \cline{2-7}   
             & Auc & AP & Auc & AP & Auc & AP   \\\hline
            GAE & $91 \pm 1.02$ & $91.9 \pm 0.89$ & $90.1 \pm 1.28$ & $92 \pm 0.96$ & $96.6 \pm 0.06$ & $96.9 \pm 0.04$  \\
            LGAE & $93.7 \pm 0.38$ & $94.5 \pm 0.26$ & $94.9 \pm 0.88$ & $95.4 \pm 0.9$ & $97.6 \pm 0.08$ & $97.7 \pm 0.12$  \\
            ARGA & $93.1 \pm 0.34$ & $93.2 \pm 0.44$ & $93.9 \pm 0.63$ & $94.3 \pm 0.65$ & $91.1 \pm 1.04$ & $90.1 \pm 1.08$  \\
            GNAE & $95 \pm 0.34$ & $95.6 \pm 0.25$ & $95.1 \pm 0.48$ & $96.1 \pm 0.4$ & $97.5 \pm 0.11$ & $97.4 \pm 0.09$  \\
            DGA & $\boldsymbol{ 99.5 \pm 0.11}$ & $\boldsymbol{99.6 \pm 0.03}$ & $\boldsymbol{99.7 \pm 0.17}$ & $\boldsymbol{99.7 \pm 0.12}$ & $\boldsymbol{99.2 \pm 0.04}$ & $\boldsymbol{99.2 \pm 0.05}$ \\\hline
            VGAE & $90.9 \pm 0.4$ & $92.5 \pm 0.28$ & $89.2 \pm 0.99$ & $91.3 \pm 0.55$ & $95.7 \pm 0.21$ & $96 \pm 0.2$  \\
            LVGAE & $94.7 \pm 0.67$ & $94.9 \pm 0.82$ & $94.9 \pm 0.36$ & $95.4 \pm 0.35$ & $97.6 \pm 0.15$ & $97.6 \pm 0.23$  \\
            ARVGA & $91.1 \pm 0.7$ & $91.2 \pm 0.74$ & $91.9 \pm 1.21$ & $92.5 \pm 1.13$ & $95.4 \pm 0.2$ & $95.9 \pm 0.1$  \\
            SIGVAE & $91.5 \pm 0.68$ & $92.6 \pm 0.71$ & $89.5 \pm 0.82$ & $91.4 \pm 0.57$ & - & -   \\
            VGNAE & $95.1 \pm 1.27$ & $95.6 \pm 1.17$ & $95.1 \pm 0.54$ & $96 \pm 0.55$ & $97.4 \pm 0.15$ & $97.4 \pm 0.14$  \\
            DVGA & $\boldsymbol{97.1 \pm 0.21}$ & $\boldsymbol{97.2 \pm 0.18}$ & $\boldsymbol{95.6 \pm 0.05}$ & $\boldsymbol{96.2 \pm 0.12}$ & $\boldsymbol{97.6 \pm 0.11}$ & $\boldsymbol{97.8 \pm 0.06}$ \\\hline
        \end{tabular}
    }}
    \label{Tab:link prediction citation network}
\end{table*}
\begin{table*}[!ht]
    \renewcommand{\arraystretch}{1}
    \caption{Results for Link Prediction of synthetic graphs with varying numbers of latent factors.($*$ denotes dataset without features, i.e., $X=\rm{I}$). Higher is better.}
    \centering
    \setlength\tabcolsep{1mm}{
     \scalebox{1}{
    \begin{tabular}{lcccccccc}
    \hline
                   \multirow{2}{*}{ \textbf{Model}} & \multicolumn{2}{c}{\textbf{4}} & \multicolumn{2}{c}{\textbf{6}} & \multicolumn{2}{c}{\textbf{8}} & \multicolumn{2}{c}{\textbf{10}} \\ \cline{2-9}
         & Auc & AP & Auc & AP & Auc & AP & Auc & AP \\ \hline
        GAE & $75.2 \pm 1.24$ & $73.2 \pm 1.18$ & $67.7 \pm 0.61$ & $66.4 \pm 0.58$ & $63.9 \pm 0.34$ & $62.2 \pm 0.43$ & $59.8 \pm 0.25$ & $57.9 \pm 0.32$ \\ 
        DGA & $\boldsymbol{80.7 \pm 1.04}$ & $\boldsymbol{81.8 \pm 0.74}$ & $\boldsymbol{79.8 \pm 1.04}$ & $\boldsymbol{81.3 \pm 1.19}$ & $\boldsymbol{79.7 \pm 0.96}$ & $\boldsymbol{80.6 \pm 0.81}$ & $\boldsymbol{76.1 \pm 1.7}$ & $\boldsymbol{79 \pm 1.63}$ \\ \hline
        VGAE & $74.8 \pm 0.74$ & $72.6 \pm 0.64$ & $58.4 \pm 2.08$ & $57.1 \pm 2.04$ & $55.9 \pm 2.26$ & $55 \pm 1.94$ & $51.6 \pm 0.63$ & $51.5 \pm 0.84$ \\ 
        DVGA & $\boldsymbol{79.6 \pm 1.61}$ & $\boldsymbol{83.6 \pm 0.89}$ & $\boldsymbol{68.3 \pm 1.94}$ & $\boldsymbol{74.5 \pm 1.11}$ & $\boldsymbol{68.5 \pm 1.43}$ & $\boldsymbol{72.7 \pm 1.74}$ & $\boldsymbol{65.9 \pm 1.87}$ & $\boldsymbol{70.5 \pm 1.1}$ \\ \hline
    \end{tabular}}}
    \label{Tab:synthetic}
\end{table*}
\textit{6) Experimental Results:} Table \ref{Tab:link prediction citation network without feature} and \ref{Tab:link prediction citation network} present the results of link prediction on three real-world citation networks without and with node features, respectively. As observed, DGA and DVGA consistently outperform other GAE-based or VGAE-based methods, which demonstrates the effectiveness of our models' structure in generating more accurate predictions. 
For instance, DVGA achieves a prediction accuracy increase of 2.1\% and 0.5\% when compared to the best-performing VGAE-based baselines on Cora and CiteSeer, respectively. Similarly, DGA demonstrates prediction accuracy improvements of 4.7\%, 4.8\%, and 1.6\% in terms of AUC against the most powerful GAE-Based baselines on Cora, CiteSeer, and PubMed, respectively. 

The formation processes of real-world graphs are typically unobservable, which makes it difficult to obtain semantic information about latent factors. In order to gain a deeper understanding and more precise analysis of our models' behavior, as mentioned earlier before, we generate synthetic datasets. We present link prediction results in Table \ref{Tab:synthetic} with varying the number of channels ($K$) for latent factors. Our results indicate that our models consistently outperform baselines when varying the number of latent factors. Particularly, as the number of factors increases from 4 to 10, the performance of GAE and VGAE steadily decreases. In contrast, our models' performance only shows a small decline, indicating that they are not affected by changes in $K$. Especially, DVGA exhibits a substantial improvement over VGAE, surpassing it by 27.3\% in AUC and 36.9\% in AP on the synthetic graph with ten latent factors. The current methods have limitations in identifying the underlying latent factors that are essential for exploring graph properties, which makes it difficult to learn disentangled representations. In contrast, our approach explicitly takes into account the entanglement of heterogeneous factors and designs a decoder specifically suited for this situation. As a result, our method demonstrates higher accuracy than existing approaches in the task of link prediction across all three datasets.
\subsection{Node clustering}
In this section, we address the task of unsupervised node clustering in the graph. We utilize the K-means clustering algorithm on the learned node embeddings, setting $K$ according to the number of classes specified in Table \ref{Tab:Statistics of citation data.}.
\begin{table*}[!ht]
    \caption{Node clustering results on Cora and CiteSeer.}
    \centering
    \begin{tabular}{lcccccccccc}
    \hline
        \multirow{2}{*}{ \textbf{Model}} & \multicolumn{5}{c}{\textbf{Cora}} & \multicolumn{5}{c}{\textbf{CiteSeer} }\\ \cline{2-11}
        ~ & acc  & precision  & F1  & NMI  & ARI  & acc & precision & F1 & NMI & ARI \\ \hline
        kmeans & $0.148$ & $0.317$ & $0.168$ & $0.183$ & $0.109$ & $0.235$ & $0.233$ & $0.206$ & $0.097$ & $0.073$ \\ 
        SC & $0.179$ & $0.146$ & $0.084$ & $0.069$ & $0.008$ & $0.182$ & $0.146$ & $0.065$ & $0.019$ & $0.001$ \\ 
        DW & $0.469$ & $0.614$ & $0.510$ & $0.413$ & $0.331$ & $0.417$ & $0.591$ & $0.439$ & $0.208$ & $0.142$ \\ \hline
        GAE & $0.286$ & $0.315$ & $0.283$ & $0.386$ & $0.318$ & $0.356$ & $0.469$ & $0.364$ & $0.125$ & $0.058$ \\ 
        VGAE & $0.400$ & $0.458$ & $0.42$ & $0.507$ & $0.483$ & $0.355$ & $0.483$ & $0.355$ & $0.189$ & $0.083$ \\
        LGAE & $0.313$ & $0.343$ & $0.325$ & $0.488$ & $0.388$ & $0.386$ & $0.479$ & $0.423$ & $0.332$ & $0.262$ \\ 
        LVGAE & $0.395$ & $0.446$ & $0.407$ & $0.500$ & $0.451$ & $0.346$ & $0.429$ & $0.359$ & $0.363$ & $0.306$ \\ 
        ARGA & $0.305$ & $0.167$ & $0.214$ & $0.400$ & $0.297$ & $0.206$ & $0.228$ & $0.213$ & $0.326$ & $0.299$ \\ 
        ARVGA & $0.281$ & $0.325$ & $0.299$ & $0.517$ & $0.475$ & $0.322$ & $0.403$ & $0.352$ & $0.287$ & $0.242$ \\ 
        GNAE & $0.197$ & $0.298$ & $0.223$ & $0.523$ & $0.475$ & $0.407$ & $0.676$ & $0.426$ & $0.266$ & $0.073$ \\ 
        VGNAE & $0.289$ & $0.329$ & $0.303$ & $0.515$ & $0.466$ & $0.293$ & $0.326$ & $0.305$ & $0.304$ & $0.255$ \\ \hline
        DGA & $0.698$ & $\boldsymbol{0.786}$ & $0.714$ & $\boldsymbol{0.526}$ & $0.479$ & $0.521$ & $0.606$ & $0.524$ & $0.295$ & $0.211$ \\ 
        DVGA & $\boldsymbol{0.738}$ & $0.766$ & $\boldsymbol{0.744}$ & $0.514$ & $\boldsymbol{0.508}$ & $\boldsymbol{0.608}$ & $\boldsymbol{0.700}$ & $\boldsymbol{0.640}$ & $\boldsymbol{0.372}$ & $\boldsymbol{0.341}$ \\ \hline
    \end{tabular}
    \label{Tab:node clustering}
\end{table*}

\textit{1) Evaluation metrics:} The node labels provided in Table \ref{Tab:Statistics of citation data.} serve as the true clustering labels for our evaluation. We adopt the approach introduced in \cite{Xia2014robust}, where we initially use the Munkres assignment algorithm \cite{Munkres1957robust} to match the predicted labels with the true labels. We employ five metrics for validating the clustering results, specifically reporting (i) accuracy (acc); (ii) precision; (iii) F1-score (F1); (iv) normalized mutual information (NMI); (v) average rand index (ARI). Consistent with previous studies such as \cite{Pan2018adversarially} and \cite{Shi2020effective}, our focus is only on Cora and CiteSeer.

\textit{2) Experimental Results:} 
Table \ref{Tab:node clustering} shows the results of node clustering for Cora and CiteSeer. The results indicate that both DGA and DVGA exhibit a substantial enhancement across all five metrics in comparison to the remaining baseline methods.
For instance, on CiteSeer, DVGA has demonstrated a 49.4\% accuracy improvement compared to GNAE and a 45.8\% improvement compared to DW. Additionally, it has achieved a 2.5\% increase in NMI compared to LVGAE and a 78.8\% increase compared to DW. On Cora, DGA has exhibited a 71.6\% precision enhancement compared to VGAE and a 28.0\% improvement compared to DW. Furthermore, it has shown a 0.6\% increase in NMI compared to GNAE and a 27.4\% improvement compared to DW. This results further prove the effectiveness and superiority of our models.
\begin{table}[!ht]
    \caption{Classification accuracies (in percent). Baseline results from \cite{Kipf2016semi}.}
    \centering
    \begin{tabular}{lccc}
        \hline
        \textbf{Model} & \textbf{Cora} & \textbf{CiteSeer} & \textbf{PubMed} \\ \hline
        ManiReg & $59.5$ & $60.1$ & $70.7$ \\ 
        SemeEmb & $59$ & $59.6$ & $71.1$ \\ 
        LP & $68$ & $45.3$ & $63$ \\ 
        DeepWalk & $67.2$ & $43.2$ & $65.3$ \\ 
        ICA & $75.1$ & $69.1$ & $73.9$ \\ 
        Planetoid & $75.7$ & $64.7$ & $77.2$ \\ 
        GCN & $81.5$ & $\boldsymbol{70.3}$ & $79$ \\ \hline
        DGA & $80.8$ & $70.2$ & $79.5$ \\ 
        DVGA & $\boldsymbol{81.9}$ & $65.6$ & $\boldsymbol{80.9}$ \\ \hline
    \end{tabular}
    \label{Tab:node classification}
\end{table}
\begin{figure}[!t]
\begin{center}
\subfloat[]{
\includegraphics[width=0.25\textwidth]{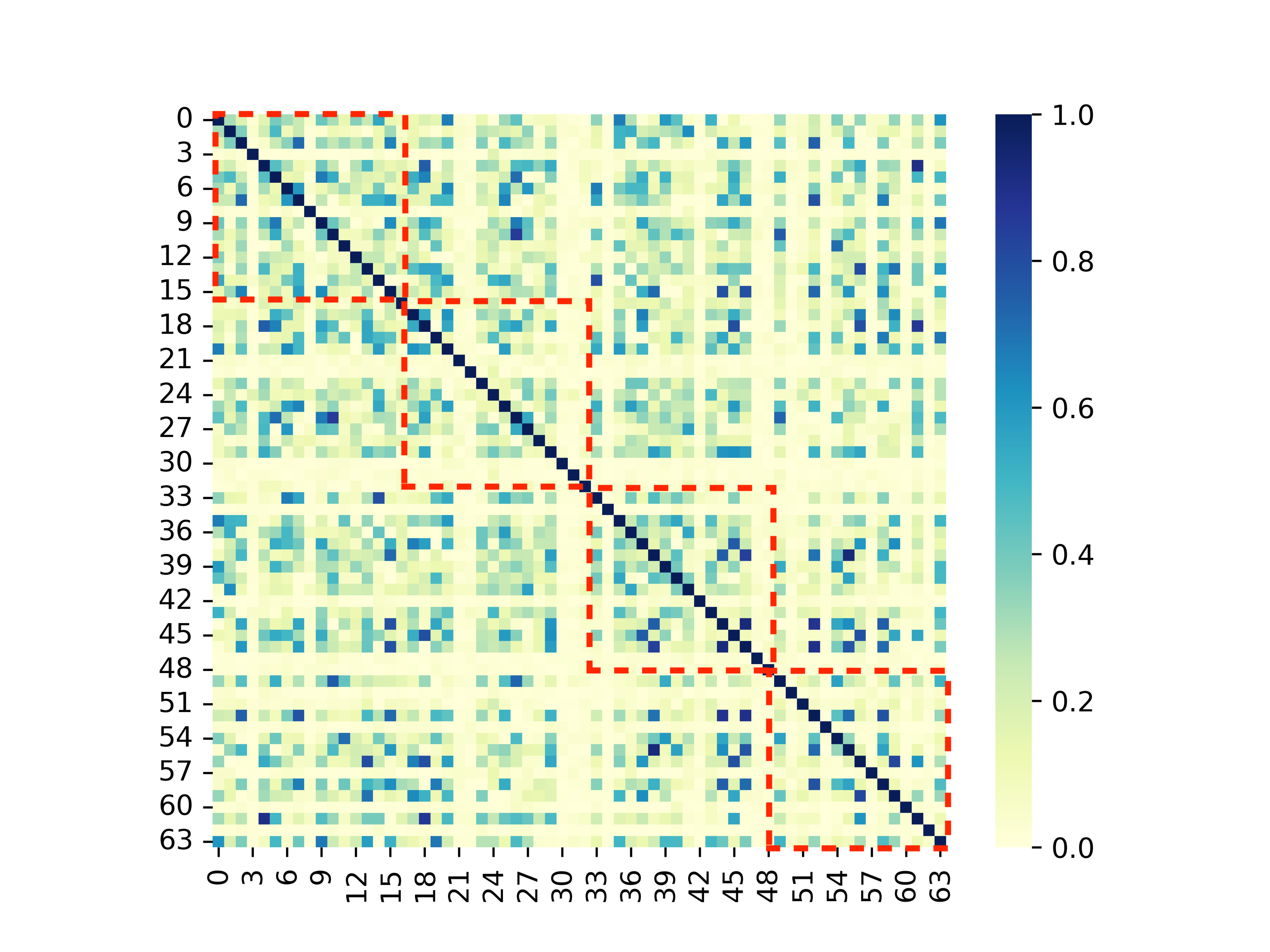}
\label{Fig:DVGE correlation}
}%
\subfloat[]{
\includegraphics[width=0.25\textwidth]{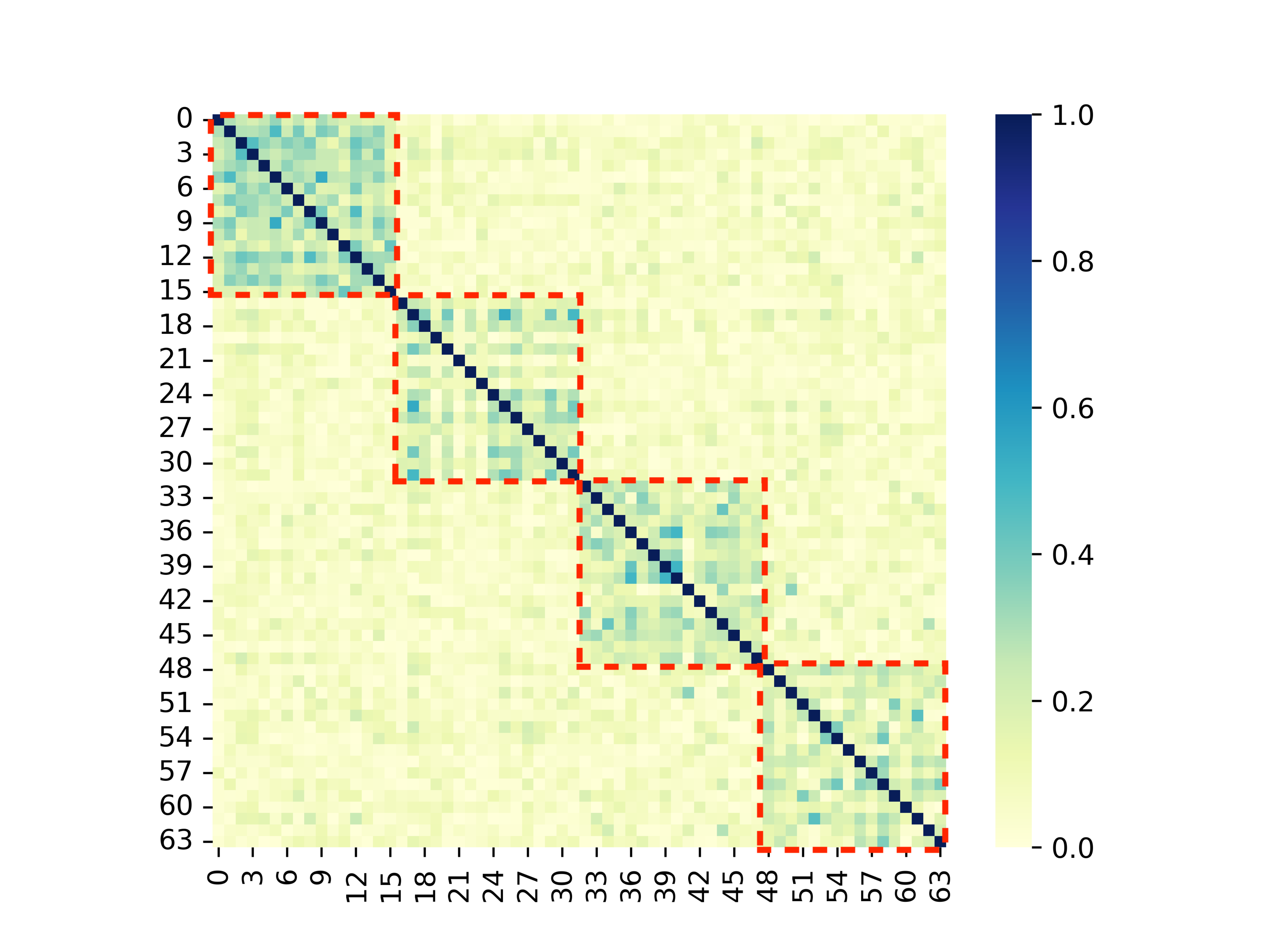}
\label{Fig:VGAE correlation}
}%
\end{center}
\caption{The absolute correlation values between the elements of the 64-dimensional representations learned by VGAE and DVGA on a synthetic graph with four latent factors. DVGA shows clearly four diagonal blocks (marked by the red dashed blocks), but VGAE does not show this correlation.}
\label{fig:correlation two}
\end{figure}
\subsection{Semi-supervised node classification}
The node classification task aims to predict the labels of the remaining nodes by utilizing the labels of a subset of nodes in the underlying graph. We adapt DGA and DVGA for node classification on three citation networks by adjusting the loss function to incorporate both the original model loss and a semi-supervised classification term. A hyper-parameter is introduced to control the relative importance of components in the combined objective function. We employ an additional fully connected neural network to map the latent representations to softmax probabilities for the required number of classes. During the optimization process, all parameters are trained jointly.

\textit{Experimental Results:} The results of node classification are summarized in Table \ref{Tab:node classification}. We observe that DVGA outperforms other competing models on Cora and PubMed. DGA also shows comparable performance to CiteSeer, where it closely aligns with the GCN, considered the best baseline. The competitive performance in comparison to state-of-the-art methods highlights the robust capability of our models, even though they are not specifically trained for this task.
\subsection{Qualitative Evaluation}
For a more comprehensive understanding of our proposed models, we conduct two qualitative experiments for a thorough examination in parallel with VGAE. These evaluations focus on the performance of disentanglement and the informativeness learned in the embeddings.

\textit{1) Correlation of Latent Features:} 
We plot the correlation between the dimensions of latent features on the synthetic graph with four latent factors in Figure \ref{fig:correlation two}. The figure illustrates the absolute correlation values between elements within the 64-dimensional graph representations obtained from VGAE and DVGA. DVGA's correlation plot exhibits a clear blockwise correlation pattern, specifically with four distinct diagonal blocks. This suggests that the four channels in DVGA are quite likely to capture mutually exclusive information. Therefore, our method successfully learns disentangled node embeddings and captures the latent factors leading to connections to some degree. This will enhance the interpretability of the link prediction results.

\textit{2) Visualization of Node Embeddings:} 
\begin{figure*}[!t]
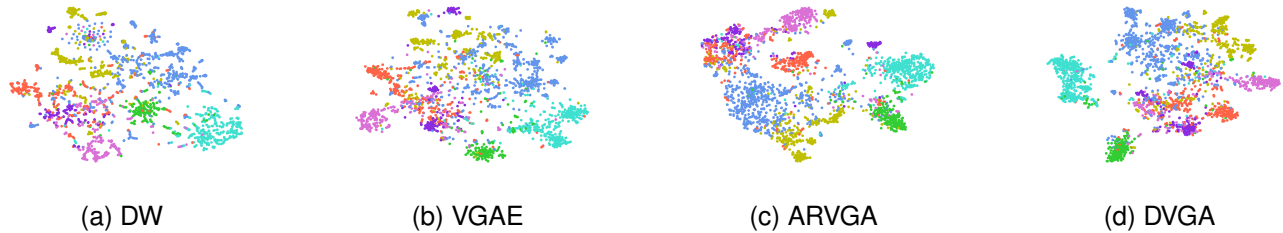

\begin{center}
\subfloat[DW]{
\includegraphics[width=0.24\textwidth]{DW_cora.pdf}
\label{Fig:DW_cora_tsne}
}%
\subfloat[VGAE]{
\includegraphics[width=0.24\textwidth]{mean_tsne_corr_200.pdf}
\label{Fig:VGAEmean_tsne}
}%
\subfloat[ARVGA]{
\includegraphics[width=0.24\textwidth]{mean-2.pdf}
\label{Fig:ARVGAmeantsne}
}%
\subfloat[DVGA]{
\includegraphics[width=0.24\textwidth]{seed-1694925034-flow-mean-03000.pdf}
\label{Fig:DVGAtsne}
}%
\end{center}
\caption{Visualization of node embeddings on Cora.}
\label{fig:tsne}
\end{figure*}
Figure \ref{fig:tsne} provides a visual comparison of the learned node embeddings from DW, VGAE, ARVGA, and DVGA on Cora. Utilizing t-SNE \cite{Van2008visualizing}, we project the node representations into a two-dimensional space. It is evident that DVGA generally learns superior node embeddings, showing high intraclass similarity and interclass differences. These findings emphasize the effectiveness of DVGA in learning disentangled representations of graph-structured data.
\subsection{Analysis}
\begin{table*}[!ht]
    \caption{Ablation Studies of DGA. }
    \centering
    \begin{tabular}{ccccccc}
    \hline
        \multirow{2}{*}{\textbf{Model}} & \multicolumn{2}{c}{\textbf{Cora}} & \multicolumn{2}{c}{\textbf{CiteSeer}
        } & \multicolumn{2}{c}{\textbf{PubMed}} \\ \cline{2-7}
         & Auc & AP & Auc & AP & Auc & AP \\ \hline
        $w/o$ indep & $99.2 \pm 0.08$ & $99.1 \pm 0.15$ & $99.3 \pm 0.1$ & $99.2 \pm 0.09$ & $98.4 \pm 0.02$ & $98.3 \pm 0.01$ \\ 
        $w/o$ disen & $98.9 \pm 0.19$ & $98.8 \pm 0.15$ & $99.0 \pm 0.21$ & $98.7 \pm 0.24$ & $99.0 \pm 0.1$ & $98.9 \pm 0.1$ \\
        DGA & $\boldsymbol{99.5 \pm 0.11}$ & $\boldsymbol{99.6 \pm 0.03}$ & $\boldsymbol{99.7 \pm 0.17}$ & $\boldsymbol{99.7 \pm 0.12}$ & $\boldsymbol{99.2 \pm 0.04}$ & $\boldsymbol{99.2 \pm 0.05}$ \\ \hline
    \end{tabular}
    \label{Tab:Ablation Studies of DGA.}
\end{table*}
\begin{table*}[!ht]
    \caption{Ablation Studies of DVGA. }
    \centering
    \begin{tabular}{ccccccc}
    \hline
        \multirow{2}{*}{\textbf{Model}} & \multicolumn{2}{c}{\textbf{Cora}} & \multicolumn{2}{c}{\textbf{CiteSeer}
        } & \multicolumn{2}{c}{\textbf{PubMed}} \\ \cline{2-7}
         & Auc & AP & Auc & AP & Auc & AP \\ \hline
        $w/o$ flow & $92.5 \pm 0.23$ & $92.7 \pm 0.78$ & $90.1 \pm 1.05$ & $91.3 \pm 0.95$ & $91.7 \pm 0.05$ & $92.7 \pm 0.14$ \\ 
        $w/o$ indep & $97.0 \pm 0.07$ & $97.2 \pm 0.09$ & $94.7 \pm 0.16$ & $95.3 \pm 0.33$ & $97.7 \pm 0.01$ & $97.9 \pm 0.02$ \\ 
        $w/o$ disen & $90.4 \pm 0.41$ & $91.1 \pm 0.41$ & $87.7 \pm 0.88$ & $89.0 \pm 0.72$ & $89.8 \pm 0.3$ & $91.2 \pm 0.34$ \\
        DVGA & $\boldsymbol{97.1 \pm 0.21}$ & $\boldsymbol{97.2 \pm 0.18}$ & $\boldsymbol{95.6 \pm 0.05}$ & $\boldsymbol{96.2 \pm 0.12}$ & $\boldsymbol{97.6 \pm 0.11}$ & $\boldsymbol{97.8 \pm 0.06}$ \\ \hline
    \end{tabular}
    \label{Tab:Ablation Studies of DVGA.}
\end{table*}
\begin{figure}[!h]
\hfill
\begin{center}
\includegraphics[width=3in]{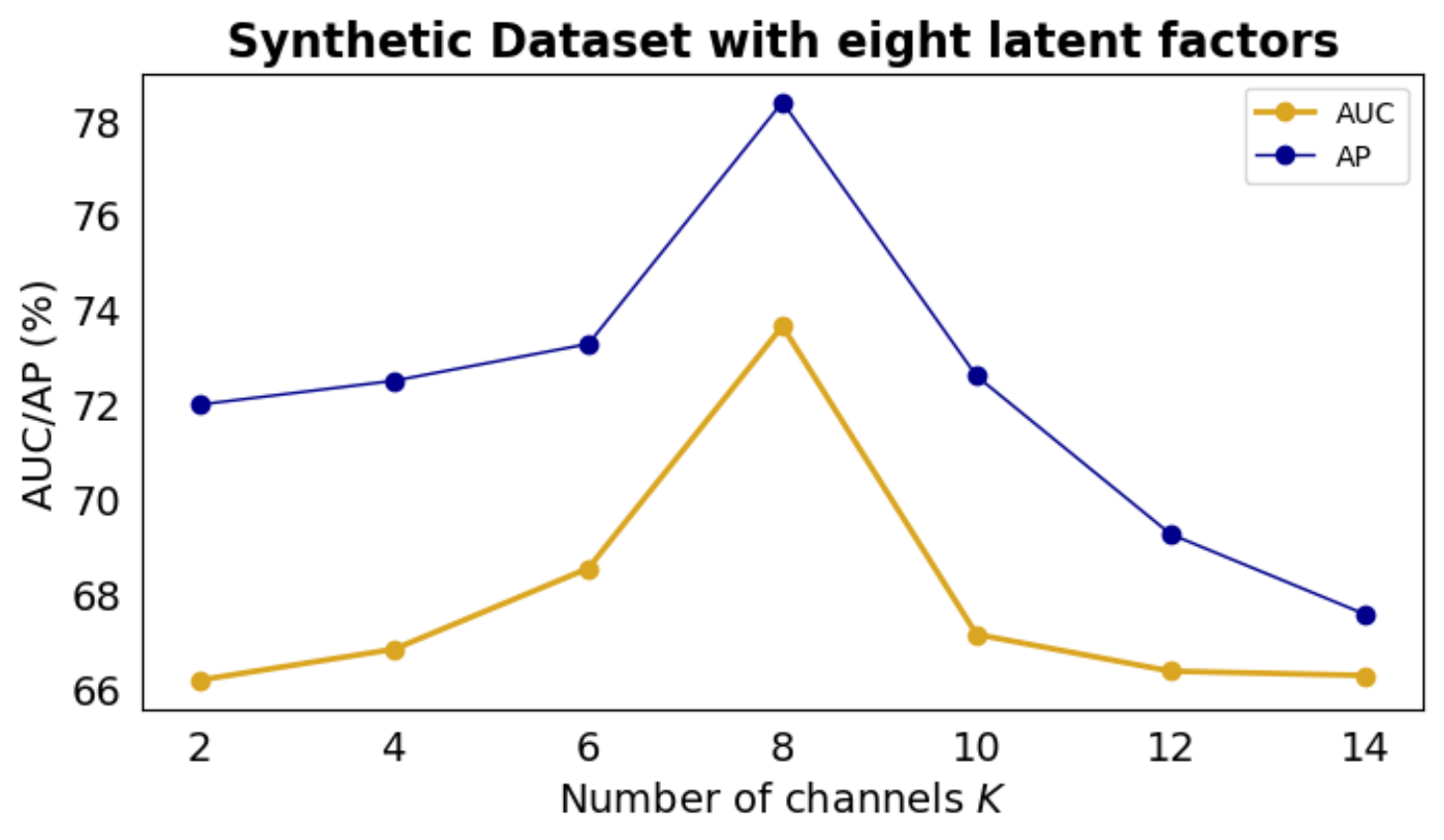}
\end{center}
\caption{Results for link prediction of synthetic graphs with different number of channels $K$.}
\label{Fig:ablation_synthetic_channel}
\end{figure}
\begin{figure*}[!t]
\vspace{-1em}  
\begin{center}
\subfloat[Number of channels $K$.]{
\includegraphics[width=0.33\textwidth,height= 1in]{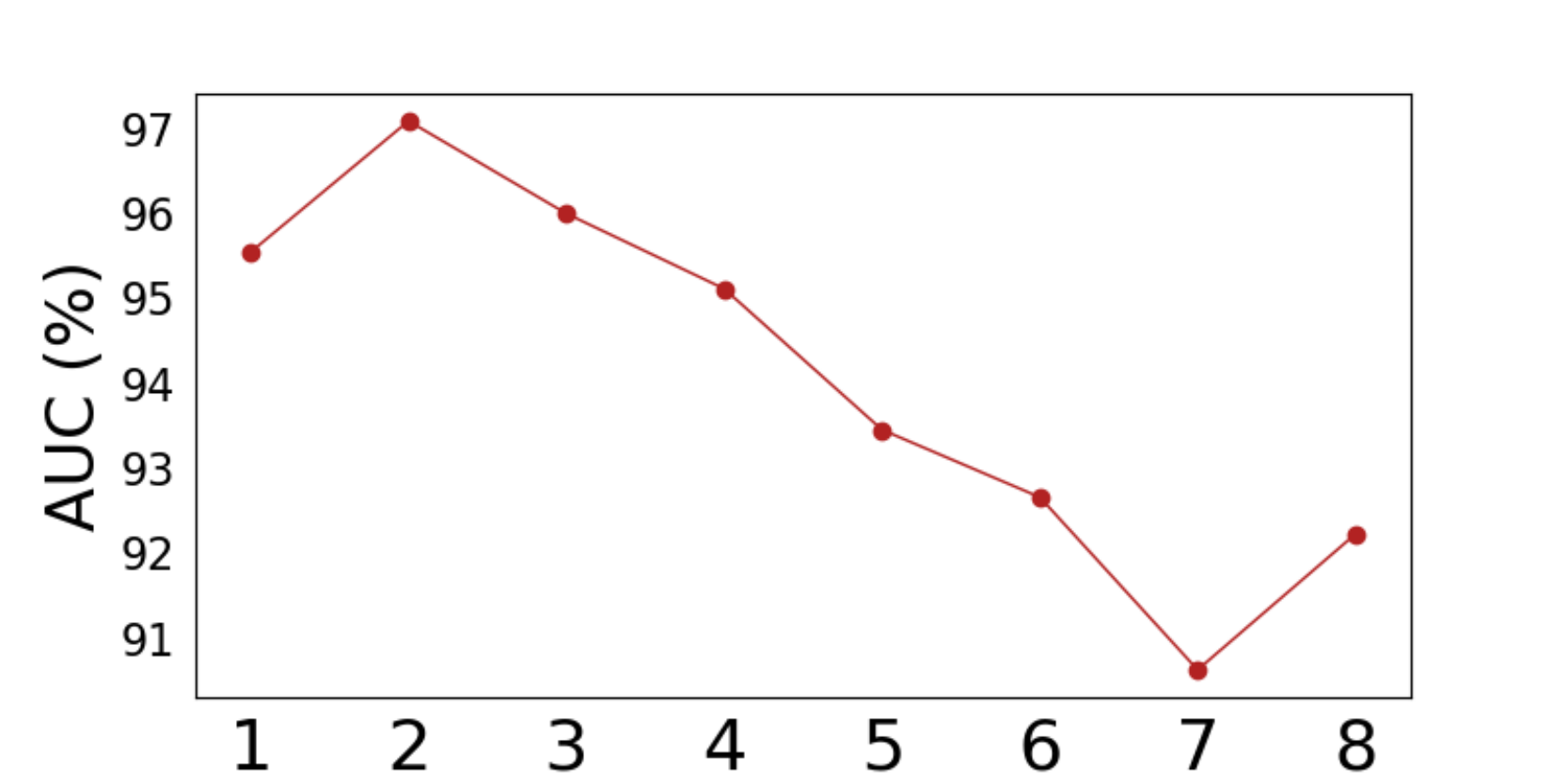}
\label{Fig:ablation_cora_channel_AUC}
}%
\subfloat[Regularization $\lambda$]{
\includegraphics[width=0.33\textwidth,height= 1in]{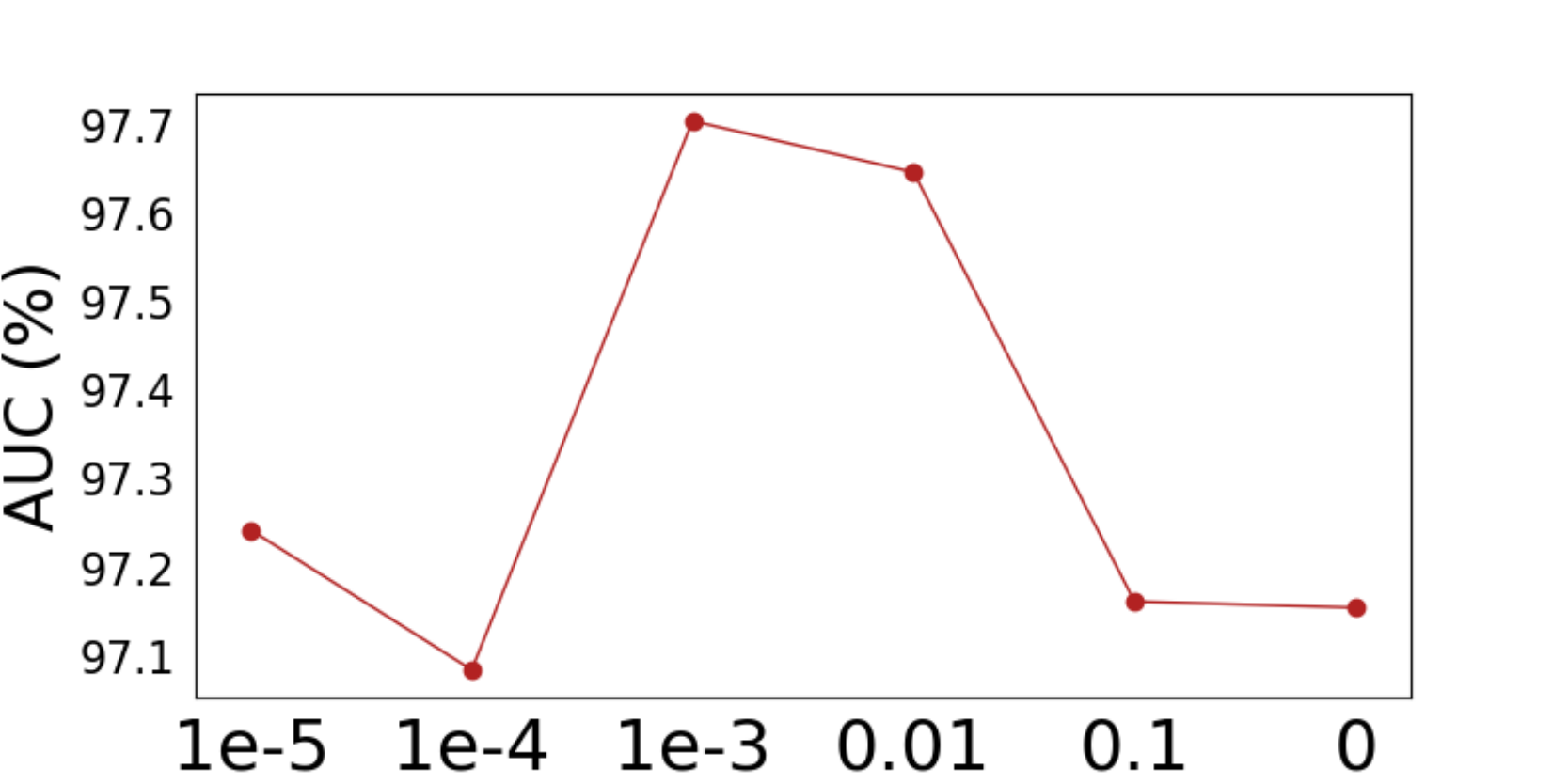}
\label{Fig:ablation_cora_lamda_AUC}
}%
\subfloat[Number of Layers $L$]{
\includegraphics[width=0.33\textwidth,height= 1in]{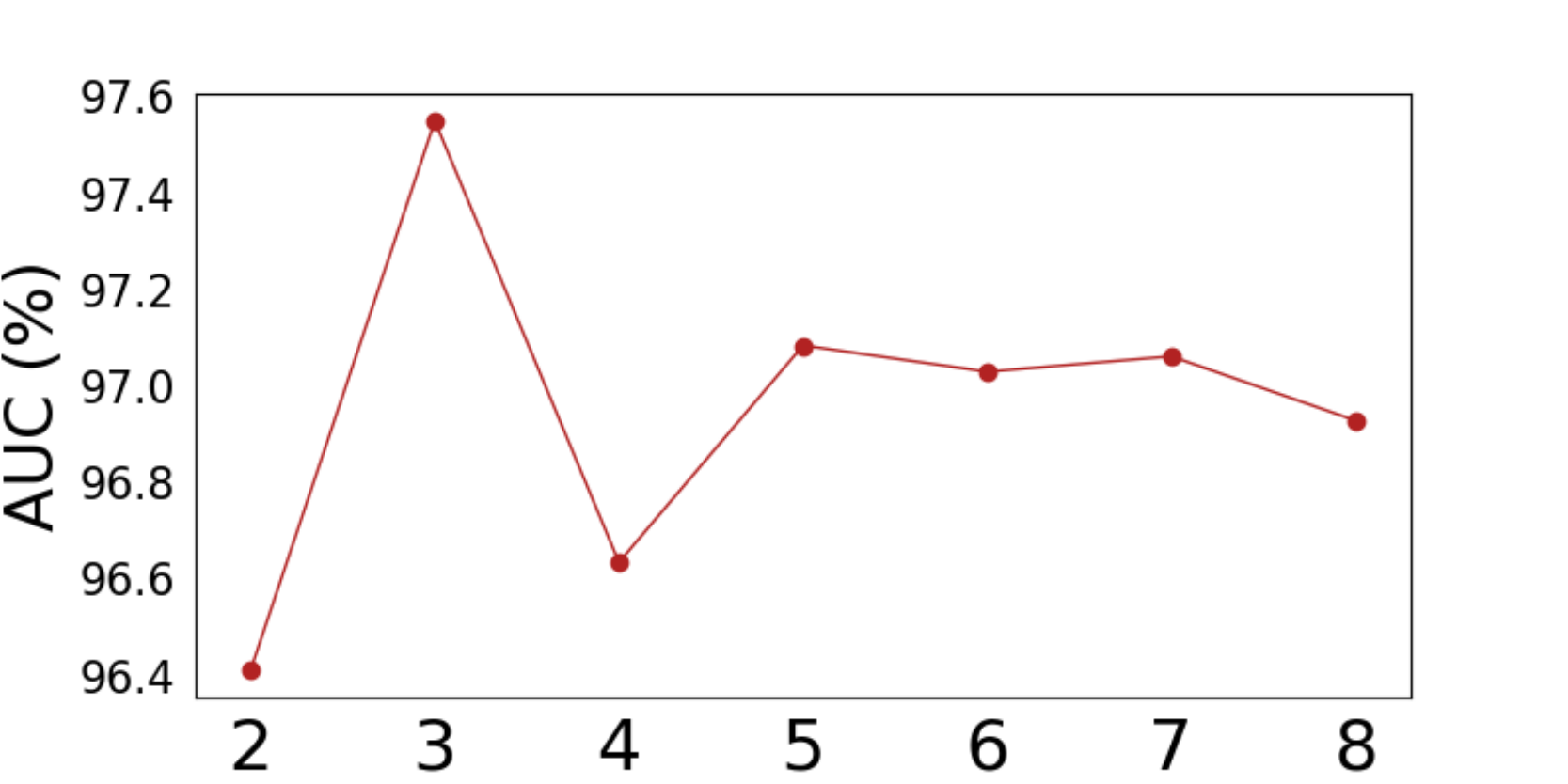}
\label{Fig:ablation_cora_layer_AUC}
}%
\end{center}
\caption{Analysis of different hyper-parameters in terms of AUC on Cora.}
\label{Fig:ablation_cora_AUC}
\vspace{-1em}  
\end{figure*}
\begin{figure*}[!t]
\vspace{-1em}
\begin{center}
\subfloat[Number of channels $K$.]{
\includegraphics[width=0.33\textwidth,height= 1in]{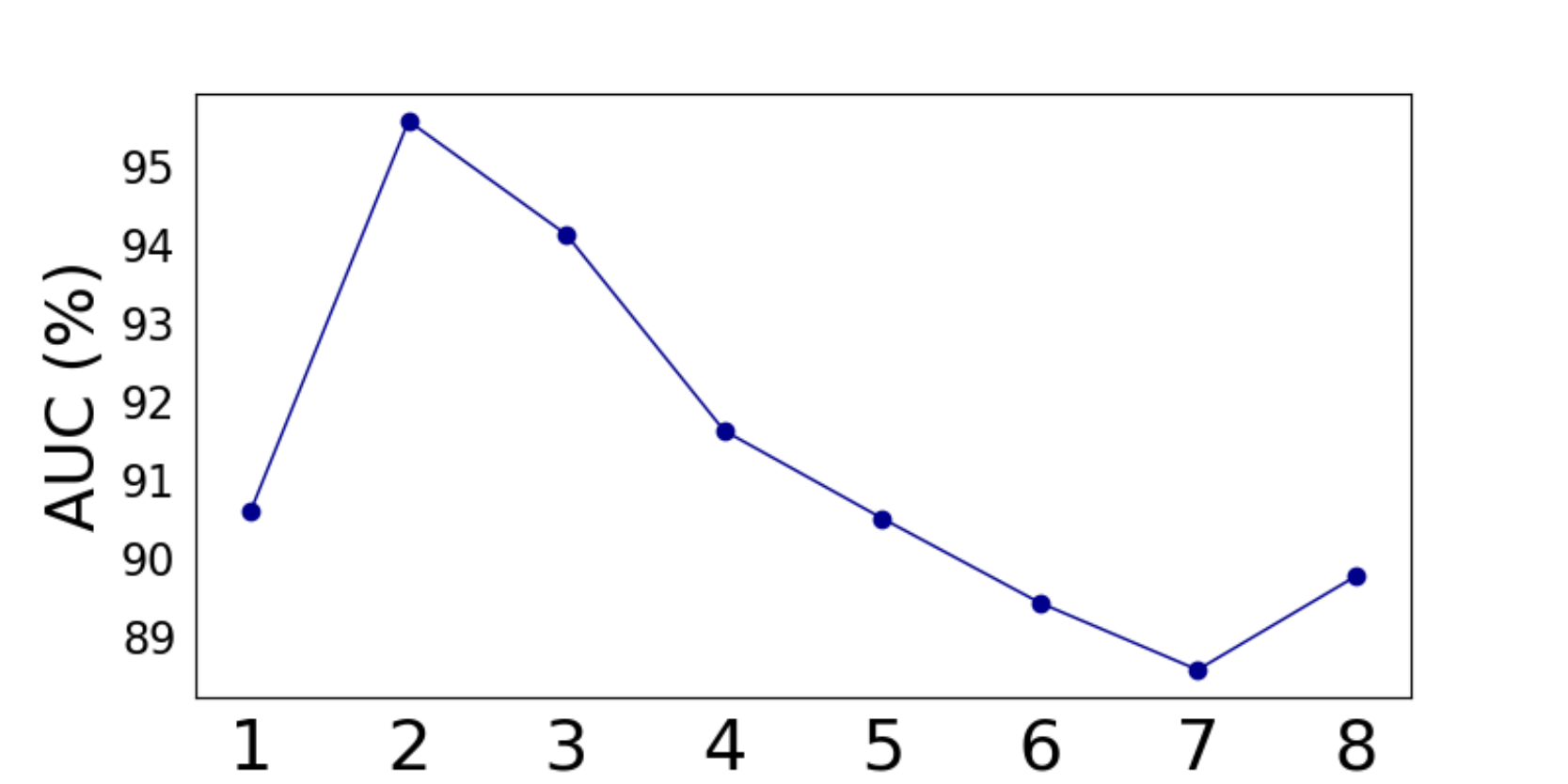}
\label{Fig:ablation_citeseer_flow_AUC}
}%
\subfloat[Regularization $\lambda$]{
\includegraphics[width=0.33\textwidth,height= 1in]{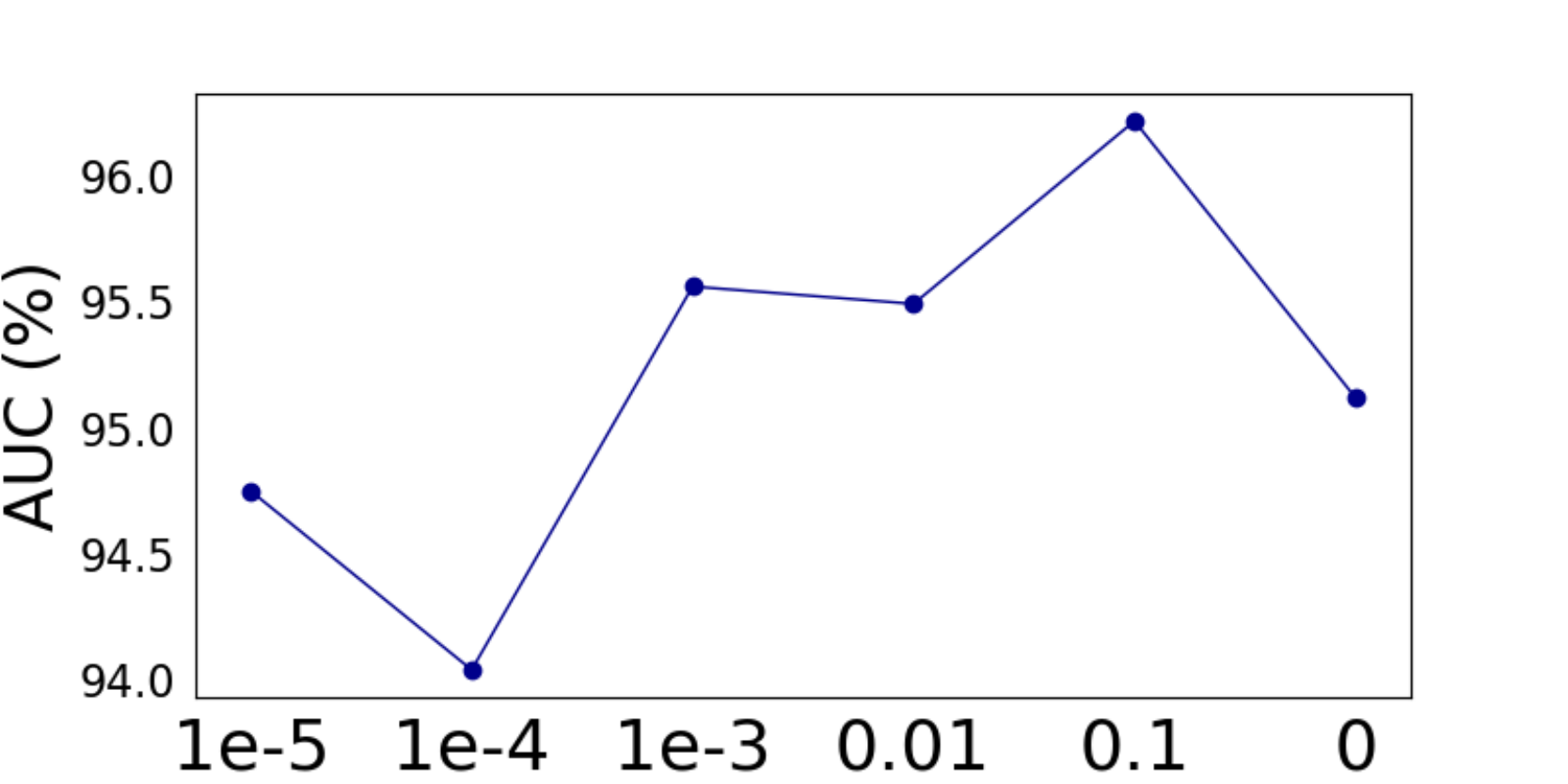}
\label{Fig:ablation_citeseer_lamda_AUC}
}%
\subfloat[Number of Layers $L$]{
\includegraphics[width=0.33\textwidth,height= 1in]{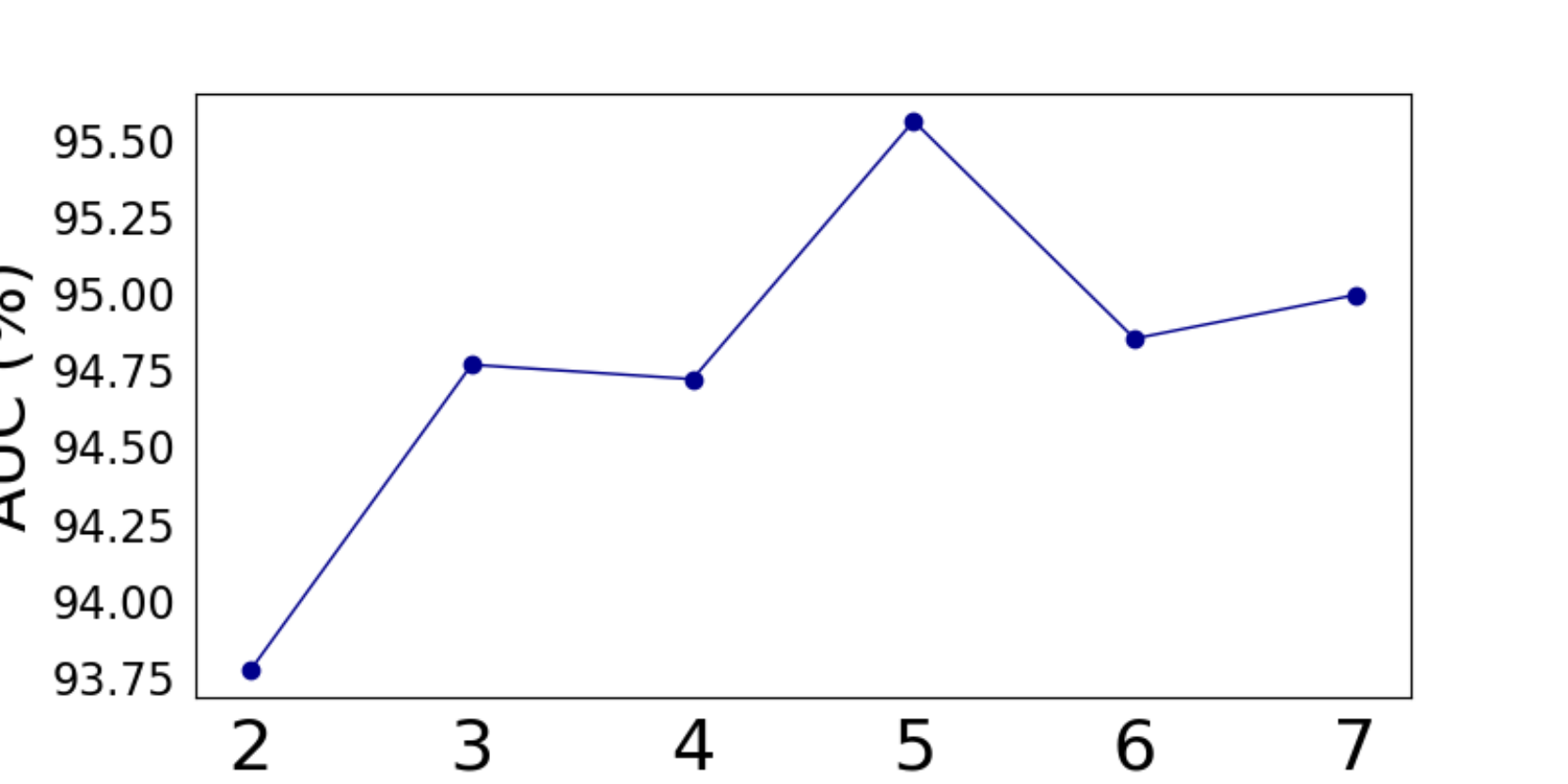}
\label{Fig:ablation_citeseer_layer_AUC}
}%
\end{center}
\caption{Analysis of different hyper-parameters in terms of AUC on CiteSeer.}
\label{Fig:ablation_citeseer_AUC}
\vspace{-1em}
\end{figure*}
\begin{figure*}[!t]
\vspace{-1em}
\begin{center}
\subfloat[Number of channels $K$.]{
\includegraphics[width=0.33\textwidth,height= 1in]{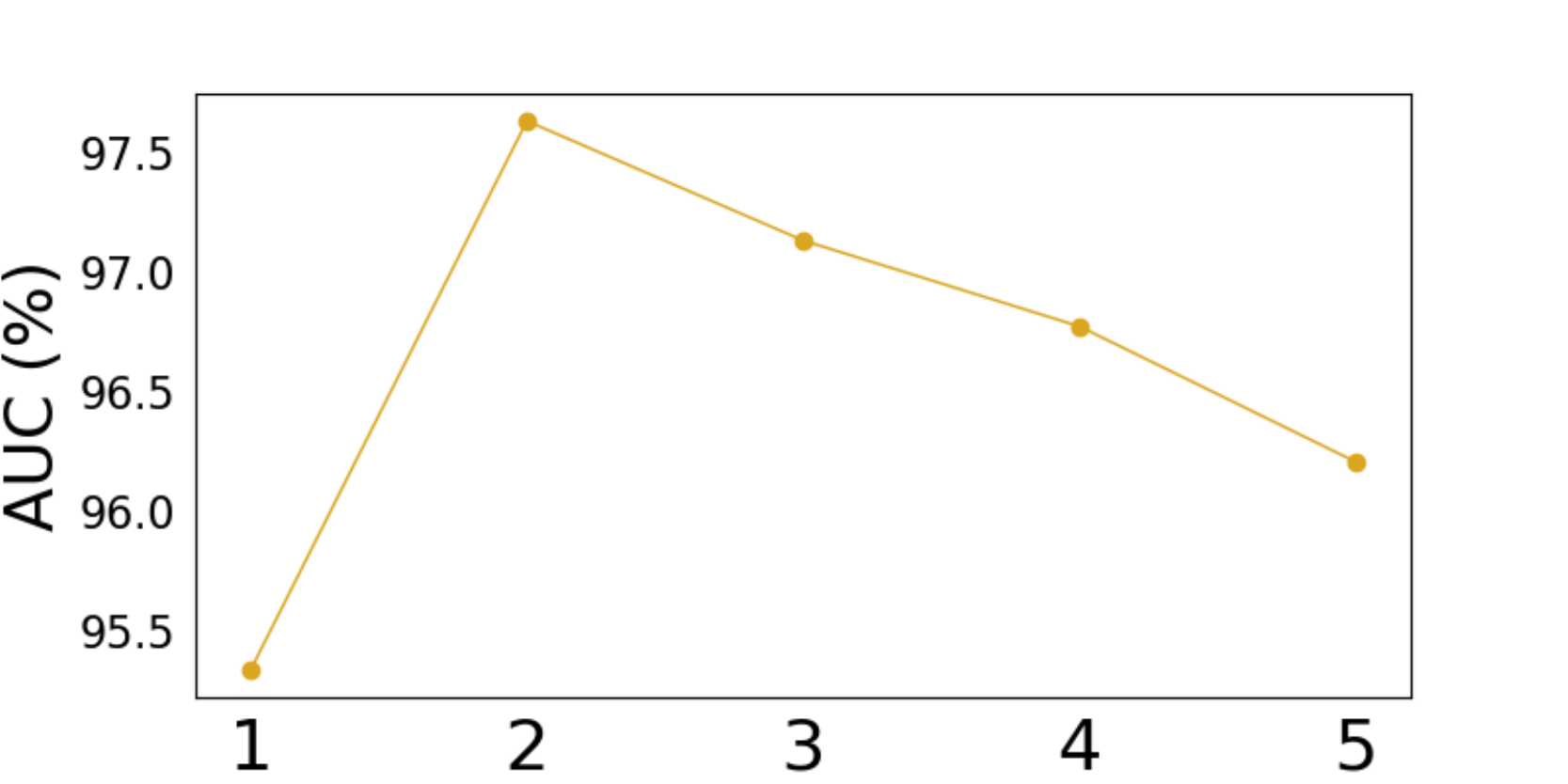}
\label{Fig:ablation_PubMed_channel_AUC}
}%
\subfloat[Regularization $\lambda$]{
\includegraphics[width=0.33\textwidth,height= 1in]{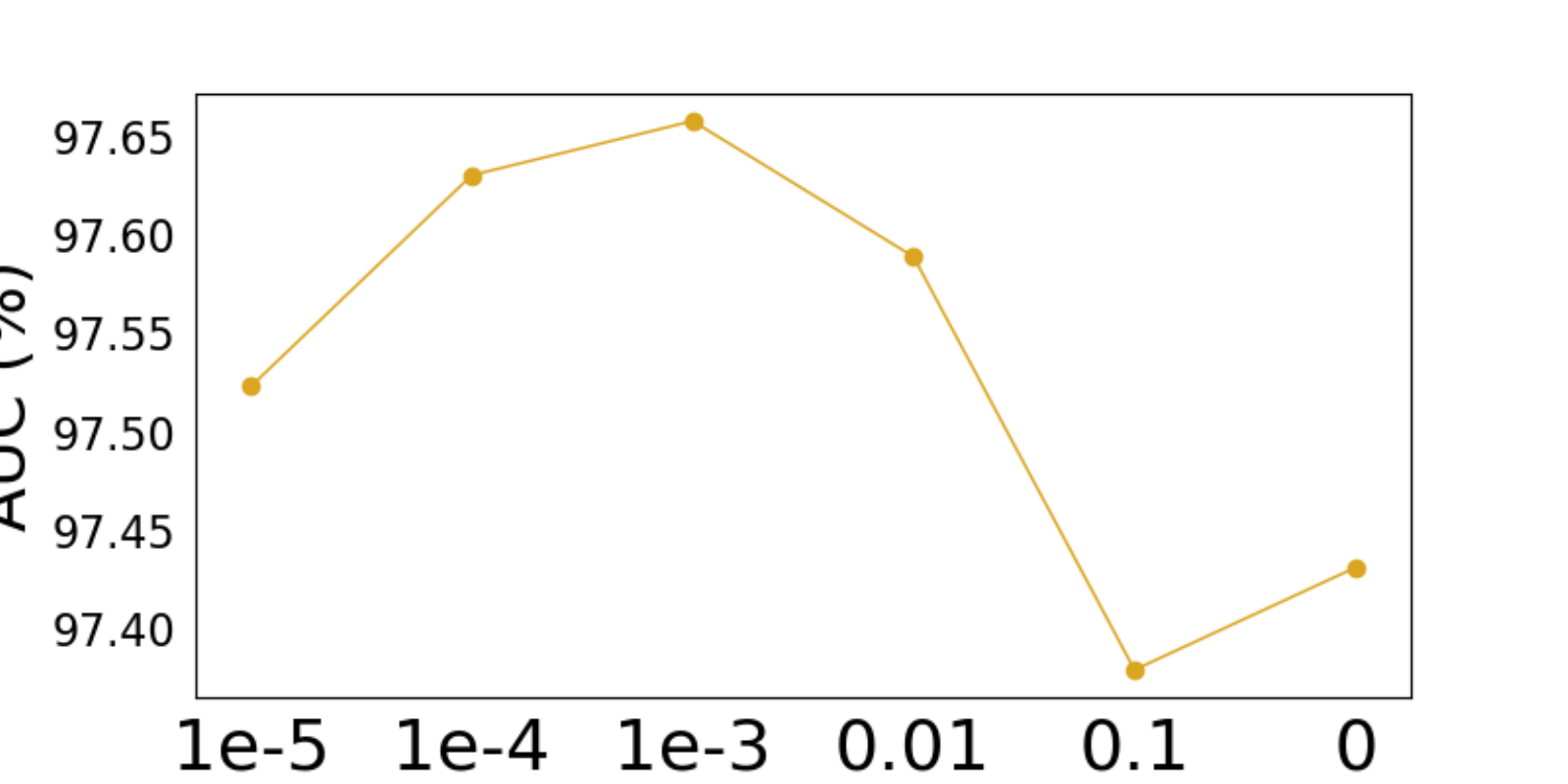}
\label{Fig:ablation_PubMed_lamda_AUC}
}%
\subfloat[Number of Layers $L$]{
\includegraphics[width=0.33\textwidth,height= 1in]{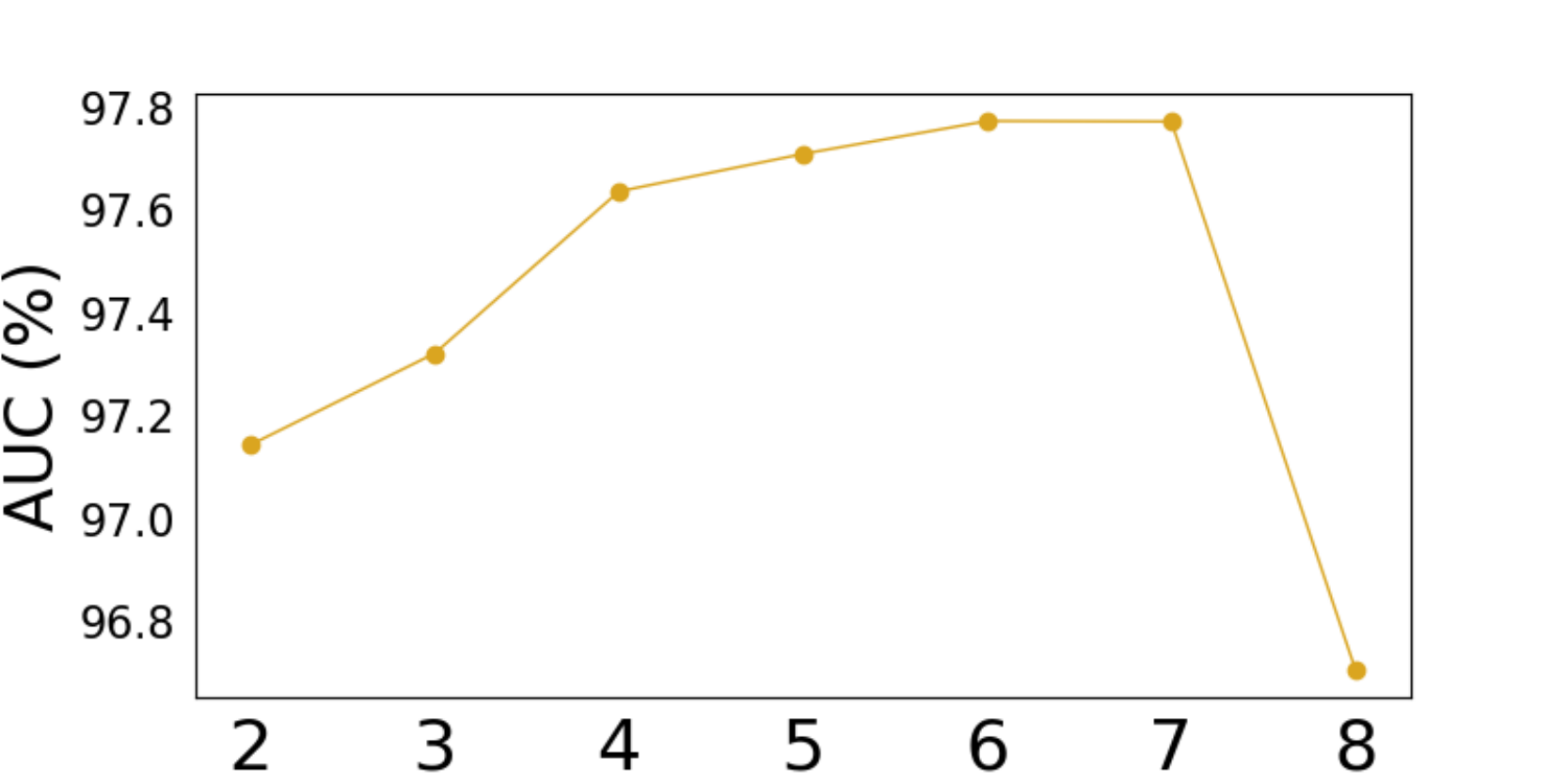}
\label{Fig:ablation_PubMed_layer_AUC}
}%
\end{center}
\caption{Analysis of different hyper-parameters in terms of AUC on PubMed.}
\label{Fig:ablation_PubMed_AUC}
\vspace{-1em}
\end{figure*}
\textit{1) Ablation Studies:} 
We conduct ablation studies on the essential components of our models to validate their contributions. We compare DVGA with the following three variants: (1) $w/o$ flow: i.e., setting $M=0$. (2) $w/o$ indep: i.e., setting $\lambda=0$. (3) $w/o$ disen: i.e., setting $K=1$, and in this case, our method degenerates into the traditional entangled variational graph auto-encoder. For DGA, as it lacks the flow component in its structure, we consider its similar variants $w/o$ indep and $w/o$ disen. For simplicity, we only report the results of the unsupervised link prediction task, with similar patterns observed in other tasks.

The results for DGA, DVGA, and their variants are presented in Tables \ref{Tab:Ablation Studies of DGA.} and \ref{Tab:Ablation Studies of DVGA.}. It is evident that removing any module diminishes the predictive capability of the models across all datasets. The elimination of the disentanglement mechanism has had a particularly large impact, resulting in a notable decrease in the performance of the model. For example, the outcomes of DVGA on Cora demonstrate this phenomenon. Therefore, we conclude that each module we designed plays a substantial role in the models' ability, and their presence collectively enhances the predictive power of the models.

\textit{2) Hyper-Parameter Sensitivity:} We investigate the sensitivity of three essential hyper-parameters. We present the evaluation results for the channel $K$ on synthetic datasets (Figure \ref{Fig:ablation_synthetic_channel}), as well as the results for all three hyperparameters on Cora (Figure \ref{Fig:ablation_cora_AUC}), CiteSeer (Figure \ref{Fig:ablation_citeseer_AUC}), and PubMed (Figure \ref{Fig:ablation_PubMed_AUC}). 
We present the AUC and AP for link prediction in terms of $K$ for DVGA on a synthetic dataset, along with the AUC for three real-world datasets in terms of all parameters. For the AP, similar results are seen on three real-world datasets, which we have eliminated for the sake of brevity.
\begin{itemize}
\item{ \textit{Number of channels $K$:} We initially examine the value of $K$ using the synthetic dataset with eight latent factors. It can be seen that the performance increases with an increment in $K$, reaching its peak at $K=8$, and then decreases. This implies that an optimal $K$ aligning with the ground-truth underlying factors outputs the best results. It also demonstrates that our model learns representations that promote disentanglement and is effective for node-related tasks. Next, we investigate the impact of channel $K$ in three citation networks. As observed previously, the performance exhibits an initial increase, followed by a peak, and then significant declines. This pattern also shows that an appropriate number of channels $K$, if matching the real number of latent factors, can lead to enhanced performance;}
\item{\textit{Regularization coefficient $\lambda$:} The regularization coefficient $\lambda$ can affect the model's performance. A large $\lambda$ tends to excessively emphasize the independence between the latent factors, whereas an excessively small $\lambda$ constrains the influence of the independence regularization. Through empirical observation, we find that setting $\lambda$ to 0.01, 0.1, and 0.001, respectively, leads to satisfactory results for Cora, CiteSeer, and PubMed.}
\item{\textit{Number of disentangle layers $L$:} The significance of the number of disentangle layers $L$ lies in the fact that a model with a small $L$ exhibits restricted capacity, potentially hindering its ability to integrate sufficient information from neighbors. From the graphs, we can observe that for Cora, the optimal 
$L$ is 3; for CiteSeer, it is 5; and for PubMed, it is 7. Additionally, when $L$ is too large, there is a decline in predictive capability, suggesting that excessive message propagation can result in overfitting, thereby compromising model performance;}
\end{itemize}
\subsection{Convergence Behavior and Complexity Analysis}\label{Sec:Convergence Behavior and Complexity Analysis}
\begin{figure}[!t]
\centering
\subfloat[]{
		\includegraphics[width=0.23\textwidth]{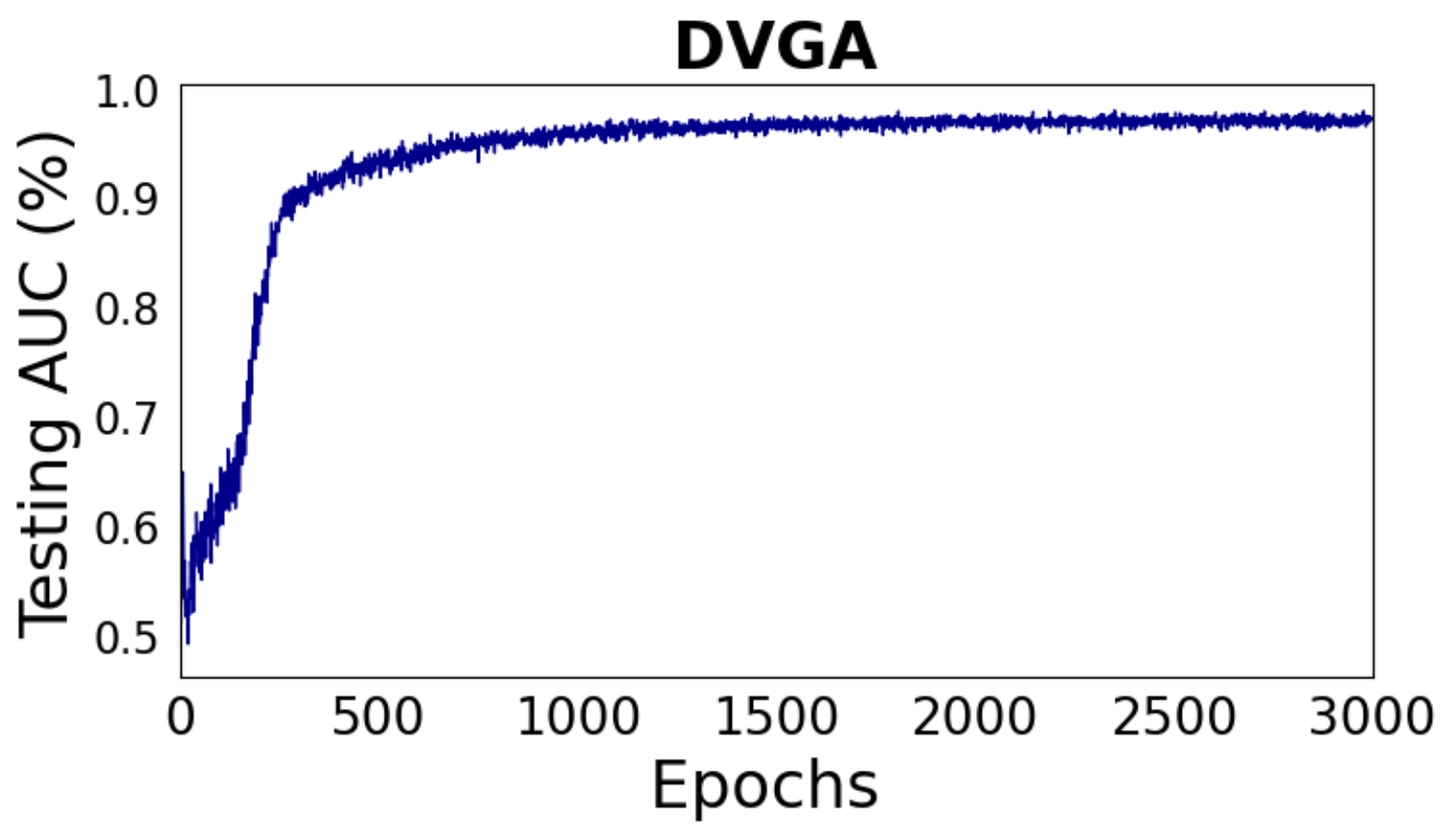}\vspace{1pt} 
  }
  \subfloat[]{
	\includegraphics[width=0.23\textwidth]{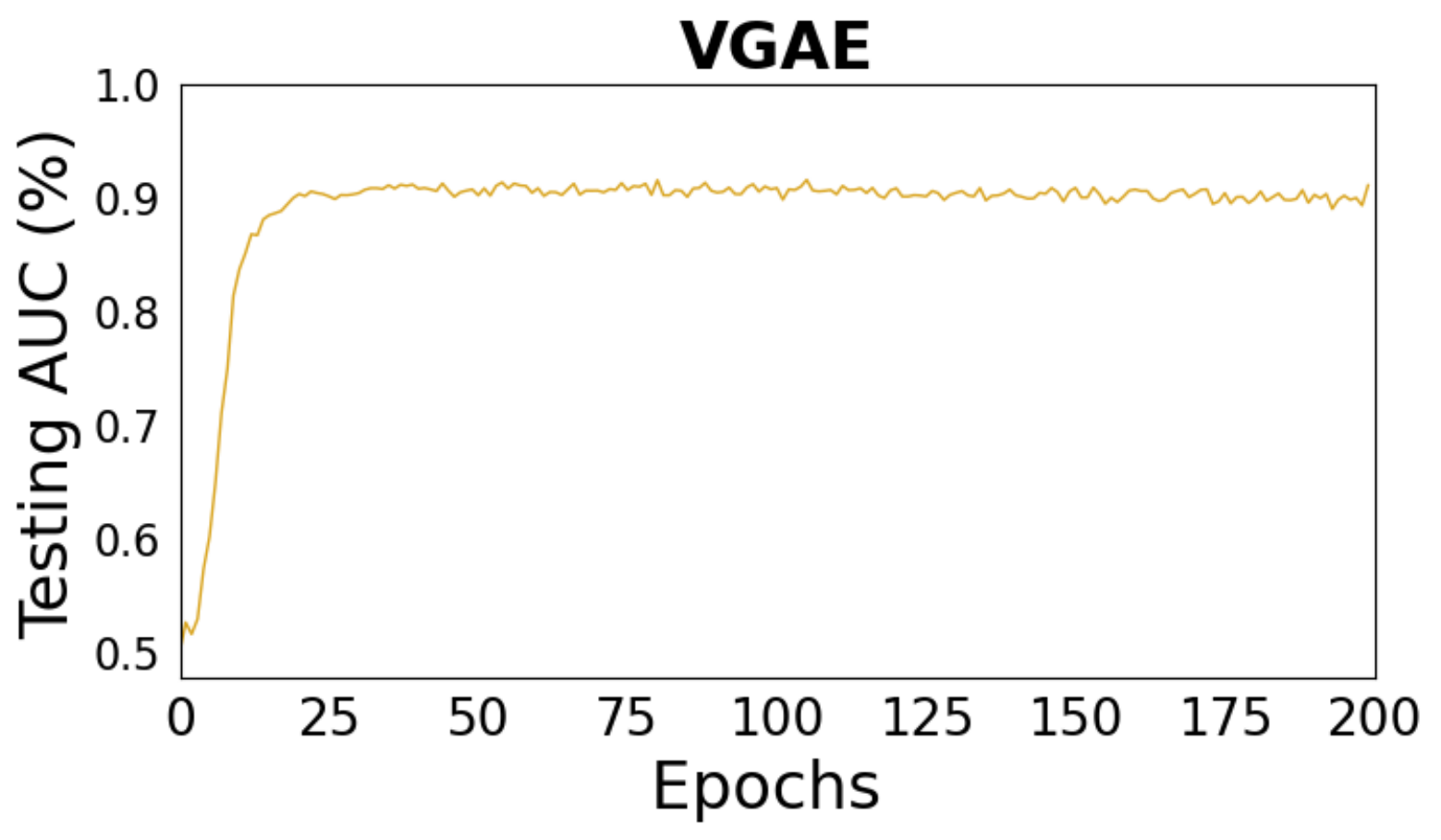}\vspace{1pt}
}
\caption{The convergence behavior of our proposed DVGA and VGAE on Cora for the link prediction task in terms of AUC. }
\label{Fig:Convergence AUC all.}
\end{figure}

Figure \ref{Fig:Convergence AUC all.} shows the evolution of testing AUC for DVGA and VGAE on Cora. Similar trends can be observed on Citeseer and PubMed. The results of testing AP exhibit a similarity to the AUC results, which we have excluded here for simplicity.
As observed in the figure, VGAE tends to easily fall into local optima, while our model exhibits greater stability, converging to higher peaks. Regarding computational complexity, Table \ref{Tab:Average training time} presents the time taken for training in each epoch. DVGA is slower on three citation network datasets compared to VGAE, mainly because of the computation of the graph with a complexity of $O(N^2)$. Nevertheless, considering the significant enhancement in performance, we believe these costs are worthwhile, particularly with the enhancement in computational capability and stability.
\begin{table}[!t]
    \caption{Average training time (s) per epoch.}
    \centering
    \begin{tabular}{lccc}
        \hline
        \textbf{Model} & \textbf{Cora} & \textbf{CiteSeer} & \textbf{PubMed} \\ \hline
        VGAE & 0.144& 0.167& 1.716 \\
        DVGA &0.296 & 0.341 & 3.635\\ \hline
    \end{tabular}
    \label{Tab:Average training time}
\end{table}
\section{Conclusion}
In this article, we investigated the problem of learning disentangled network representations using the (variational) graph auto-encoder and proposed two models, namely Disentangled Graph Auto-Encoder (DGA) and Disentangled Variational Graph Auto-Encoder (DVGA). We developed a dynamic disentangled graph encoder that can effectively aggregate features in a disentangled manner. The utilization of component-wise flow in the DVGA further facilitates the learning of expressive representations. In order to better capture distinct information between different channels in latent representations, we implemented independent regularization to promote statistical independence. In addition, a factor-wise decoder tailored for disentangled representations was built to improve the models' performance. Extensive experimental results on synthetic and real-world datasets demonstrated that our method outperforms recent baselines and effectively learns interpretable disentangled representations. As for future work, it would be worthwhile and intriguing to explore new applications that make use of the strength provided by disentangled representations.

\bibliographystyle{IEEEtran}
\bibliography{reference}

\end{document}